\pdfoutput=1
\documentclass[journal]{IEEEtran}
\usepackage{acronym}
\usepackage{siunitx}
\usepackage{graphicx}
\usepackage{tikz}
\usepackage{amsmath,amssymb}
\usepackage{multirow}
\usepackage{colortbl}

\usepackage{bm}
\usepackage[caption=false]{subfig}

\usepackage{textgreek}
\usepackage{float}
\usepackage{braket}

\acrodef{EO}{Earth Observation}
\acrodef{ESA}{European Space Agency}
\acrodef{CNN}{Convolutional Neural Network}
\acrodef{EO}{Earth Observation}
\acrodef{RS}{Remote Sensing}
\acrodef{DL}{Deep Learning}
\acrodef{ML}{Machine Learning}
\acrodef{HPC}{High-Performance Computing}
\acrodef{DEEP-EST}{Dynamic Exascale Entry Platform - Extreme Scale Technologies}
\acrodef{GPU}{Graphics Processing Unit}
\acrodef{JUWELS}{J\"{u}lich Wizard for European Leadership Science}
\acrodef{JSC}{J\"{u}lich Supercomputing Centre}
\acrodef{NCCL}{NVIDIA Collective Communication Library}
\acrodef{RGB}{Red-Green-Blue}
\acrodef{IR}{Infrared}
\acrodef{MPI}{Message Passing Interface}
\acrodef{FLOPS}{Floating Point Operations per Second}
\acrodef{HLS}{Harmonized Landsat and Sentinel}
\acrodef{GAN}{Generative Adversarial Network}
\acrodef{MGAN}{multispectral Generative Adversarial Network}
\acrodef{ESRGAN}{Enhanced Super-Resolution Generative Adversarial Network}
\acrodef{SAM}{Spectral Angle Mapper}
\acrodef{SAR}{Synthetic Aperture Radar}
\acrodef{RMSE}{Root Mean Square Error}
\acrodef{UIQI}{Universal Image Quality Index}
\acrodef{ERGAS}{Relative Dimensionless Global Error}
\acrodef{PSNR}{Peak Signal-to-Noise Ratio}
\acrodef{NAS}{Neural Architecture Search}
\acrodef{TSs}{Time Series}
\acrodef{OA}{Overall Accuracy}
\acrodef{AI}{Artificial Intelligence}
\acrodef{MSI}{Multi-Spectral Instrument}
\acrodef{NASA}{National Aeronautics and Space Administration}
\acrodef{USGS}{United States Geological Survey}
\acrodef{SPOT}{Satellite Pour l'Observation de la Terre}

\acrodef{QC}{Quantum Computing}
\acrodef{AQC}{Adiabatic Quantum Computation}
\acrodef{SVM}{Support Vector Machine}
\acrodef{MSVM}{Multiclass SVM}
\acrodef{QA}{Quantum Annealing}
\acrodef{QML}{Quantum Machine Learning}
\acrodef{OVO}{one-versus-one}
\acrodef{OVA}{one-versus-all}
\acrodef{DAG}{Directed Acyclic Graph}
\acrodef{CS}{Crammer-Singer}
\acrodef{QSVM}{Quantum SVM}
\acrodef{QMSVM}{Quantum Multiclass SVM}
\acrodef{QUBO}{Quadratic Unconstrained Binary Optimization}

\DeclareSIUnit\pixel{px}

\newcommand{\comment}[1]{}

\newcommand{\Aperta}{\Bigg( \Bigg.}
\newcommand{\Chiusa}{\Bigg. \Bigg)}
\newcommand\figurespace{0.45\linewidth}
\newcommand\figurewidth{3.5cm}

\usepackage{url}
\usepackage{algorithm}
\usepackage{algorithmic}
\makeatletter
\newcommand\fs@norules{\def\@fs@cfont{\bfseries}\let\@fs@capt\floatc@ruled
  \def\@fs@pre{}%
  \def\@fs@post{}%
  \def\@fs@mid{\kern3pt}%
  \let\@fs@iftopcapt\iftrue}
\makeatother
\floatstyle{norules}
\restylefloat{algorithm}

\begin{document}

\title{A Single-Step Multiclass SVM based on Quantum Annealing for Remote Sensing Data Classification}

\author{Amer~Delilbasic,~\IEEEmembership{Student Member,~IEEE,} 
        Bertrand Le Saux,~\IEEEmembership{Senior Member,~IEEE,}
        Morris~Riedel,~\IEEEmembership{Member,~IEEE,}
        Kristel~Michielsen,
        and~Gabriele~Cavallaro,~\IEEEmembership{Member,~IEEE}
\thanks{Amer Delilbasic is with the J\"{u}lich Supercomputing Centre, Wilhelm-Johnen Stra\ss e, 52428 J\"{u}lich, Germany, with the University of Iceland, 107 Reykjavik, Iceland, and with ESA/ESRIN $\Phi$-lab, IT-00044 Frascati, Italy (e-mail: a.delilbasic@fz-juelich.de).

Bertrand Le Saux is with ESA/ESRIN $\Phi$-lab, IT-00044 Frascati, Italy (e-mail: bertrand.le.saux@esa.int).

Morris Riedel and Gabriele Cavallaro are with the University of Iceland, 107 Reykjavik, Iceland, and the J\"{u}lich Supercomputing Centre, Wilhelm-Johnen Stra\ss e, 52428 J\"{u}lich, Germany (e-mail: morris@hi.is, g.cavallaro@fz-juelich.de).

Kristel Michielsen is with the J\"{u}lich Supercomputing Centre, Wilhelm-Johnen Stra\ss e, 52428 J\"{u}lich, Germany, and the RWTH Aachen University, D-52056 Aachen, Germany (e-mail: k.michielsen@fz-juelich.de). 

\vspace{0.5cm}
This work has been submitted to the IEEE for possible publication. Copyright may be transferred without notice, after which this version may no longer be accessible.}}
\maketitle

\begin{abstract}

In recent years, the development of quantum annealers has enabled experimental demonstrations and has increased research interest in applications of quantum annealing, such as in quantum machine learning and in particular for the popular quantum SVM.
Several versions of the quantum SVM have been proposed, and quantum annealing has been shown to be effective in them. Extensions to multiclass problems have also been made, which consist of an ensemble of multiple binary classifiers.
This work proposes a novel quantum SVM formulation for direct multiclass classification based on quantum annealing, called \ac{QMSVM}. The multiclass classification problem is formulated as a single \ac{QUBO} problem solved with quantum annealing. The main objective of this work is to evaluate the feasibility, accuracy, and time performance of this approach. Experiments have been performed on the D-Wave Advantage quantum annealer for a classification problem on remote sensing data. The results indicate that, despite the memory demands of the quantum annealer, \ac{QMSVM} can achieve accuracy that is comparable to standard SVM methods and, more importantly, it scales much more efficiently with the number of training examples, resulting in nearly constant time.
This work shows an approach for bringing together classical and quantum computation, solving practical problems in remote sensing with current hardware.

\end{abstract}


%
\IEEEpeerreviewmaketitle

\section{Introduction}

\IEEEPARstart{I}{n} the context of \acf{EO} \cite{Chi2016}, there is a growing availability of data acquired by heterogeneous \ac{RS} sources. Many applications are currently benefitting from \ac{RS} data, e.g., agriculture, green energy development and urban monitoring. The devices and software for data processing have to match this trend in order to extract information from the collected data in a timely manner.

\acf{QC} \cite{Nielsen2010}, a computational paradigm based on the postulates and laws of quantum mechanics, has proved the potential to reach an exponential algorithmic speedup with respect to classical computation under certain assumptions \cite{Jozsa2003,Ronnow2014}. Among the quantum computational models defined in the literature, two broadly employed models can be identified. The quantum circuit model \cite{Aharonov1998}, similarly to the classical circuit model, is based on circuits, gates and measurements applied to qubits (quantum bits). \acf{AQC} \cite{McGeoch2014,Albash2018} aims at solving optimization problems by exploiting the time evolution of a quantum system satisfying the requirements of the adiabatic theorem \cite{Born1928}. Despite their differences, the two models have been proven to be computationally equivalent \cite{Aharonov2008}. The focus of this work is \acf{QA} \cite{Kadowaki1998,Finnila1994}, a heuristic search approach based on \ac{AQC}, since commercially ready quantum annealers are available for analyzing the disruptive potential of \ac{QC}.

\acf{QML} \cite{Biamonte2017,Dunjko2018} is a research area working on \ac{QC} algorithms applied to \ac{ML} tasks, with the purpose of obtaining a computational speedup or a prediction accuracy increase.
\ac{QML} methods based on \ac{QA} have proven to outperform classical \ac{ML} in selected applications with limited training examples, for example in computational biology \cite{Li2018}. 
Recent studies have analyzed how \ac{QML} can be integrated into \ac{EO} tasks. In \cite{Gawron2020} and \cite{Sebastianelli2022}, circuit-based quantum neural networks have been trained for multispectral image classification.
The work of Otgonbaatar and Datcu has covered different aspects of circuit-based \ac{QML} for \ac{EO}, e.g., natural data embedding \cite{Otgonbaatar2021b}, parameterized quantum gates \cite{Otgonbaatar2022a} and transfer learning \cite{Otgonbaatar2022b}.
Circuit-based quantum kernels have been applied to binary \cite{Shaik2022} and multiclass \cite{Pai2022} \ac{RS} image classification.
\ac{QA} has also found a place in \ac{EO} for solving specific optimization problems. In \ac{SAR} imaging, problems related to system design \cite{Huber2022} and phase ambiguity \cite{Otgonbaatar2021c} have been addressed. In the context of \ac{QML}, a feature selection method for hyperspectral images has been proposed \cite{Otgonbaatar2021a}, and a \ac{QA}-based \ac{QSVM} method has been successfully used for binary classification of multispectral images \cite{Cavallaro2020} \cite{Delilbasic2021}.

The \ac{SVM} is an efficient and theoretically-grounded algorithmic approach in statistical learning theory. Different versions and formulations of the SVM can be found in the literature for a variety of tasks and applications, e.g., pattern recognition, computer vision, image analysis and business intelligence \cite{Ma2014}.
\ac{SVM} has also been proven to be effective in \ac{EO} pixel-wise image classification \cite{Melgani2004}.

Defining a SVM framework for multiclass classification is a non-trivial task. Two different approaches can be followed \cite{Hsu2002}. The {\textit{multiple-step} (or \textit{indirect}) \textit{approach} reframes the problem by defining an ensemble of multiple binary SVM classifiers and multiple classification outputs. The most common methods are the \ac{OVO}, \ac{OVA} and the \ac{DAG} \ac{SVM}. In the \ac{OVO} method, for example, each pair of classes defines a SVM classifier. The outcomes of each classifier are usually combined with a ``max wins" strategy which determines the final prediction. The \ac{QSVM} algorithms for multiclass classification available in the literature, e.g., \cite{Bishwas2018,Dema2020,Yuan2022}, follow this approach. They are defined as ensembles of binary \ac{QSVM} classifiers, which can be full quantum \cite{Rebentrost2014,Zhang2022}, quantum kernel \cite{Havlicek2019} and \ac{QA}-based \cite{Willsch2019} formulations.

The {\textit{single-step} (or \textit{direct}) \textit{approach} for multiclass classification defines a single optimization step on the whole training set, which finds boundaries between all classes in one pass.
The \ac{CS} \ac{SVM}, proposed in \cite{Crammer2002a}, is an example of single-step approach. A limitation of this method is the complexity of the training phase, due to the high number of optimization variables, which makes this approach impractical. In the same work, simplified formulations are proposed, which reduce the problem size and enable better performances, although at the cost of optimality.

The main objective of this work is to propose a novel approach, specifically \acf{QMSVM}, by reframing the original formulation of the \ac{CS} \ac{SVM}, thus enabling the optimization step to be performed using a \ac{QA} algorithm.
This work studies the computational capability offered by the available quantum annealers and test the feasibility of a \ac{QA}-based \ac{SVM} single-step approach.
Experiments are performed on a real quantum annealer, i.e., D-Wave Advantage \cite{D-Wave,McGeoch2021}, in order to validate the algorithm and analyze the potential of current hardware. The performance is evaluated both in terms of accuracy and execution time, both relevant aspects in practice.
The code repository of the algorithm is made available for reproducibility\footnote{\url{https://gitlab.jsc.fz-juelich.de/sdlrs/qmsvm}}.

\renewcommand\figurespace{0.3\linewidth}
\renewcommand\figurewidth{0.4\linewidth}
\begin{figure*}
    \centering
    \subfloat[QUBO graph]{\includegraphics[width=\figurewidth]{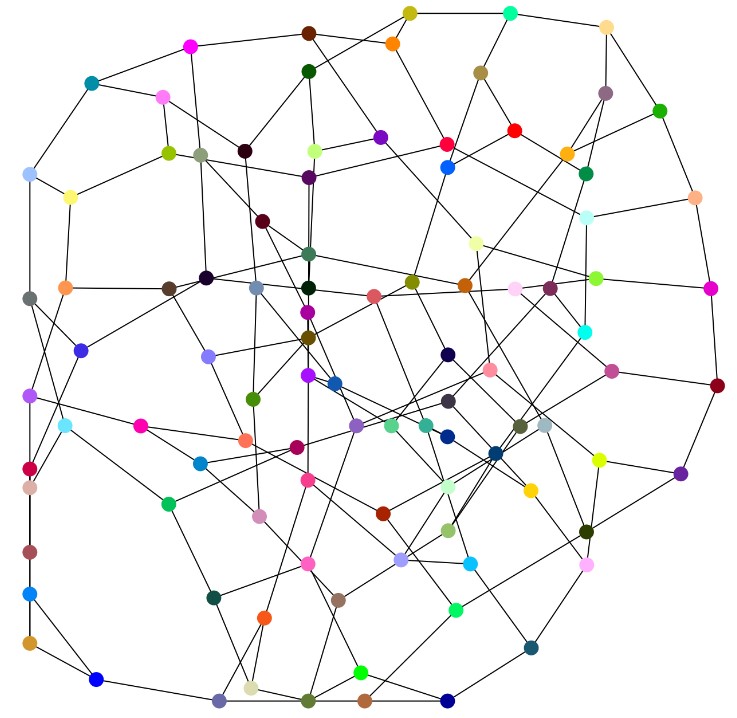}}%
    \hfil
    \subfloat[Embedding of the QUBO graph on a Pegasus architecture]{\includegraphics[width=\figurewidth]{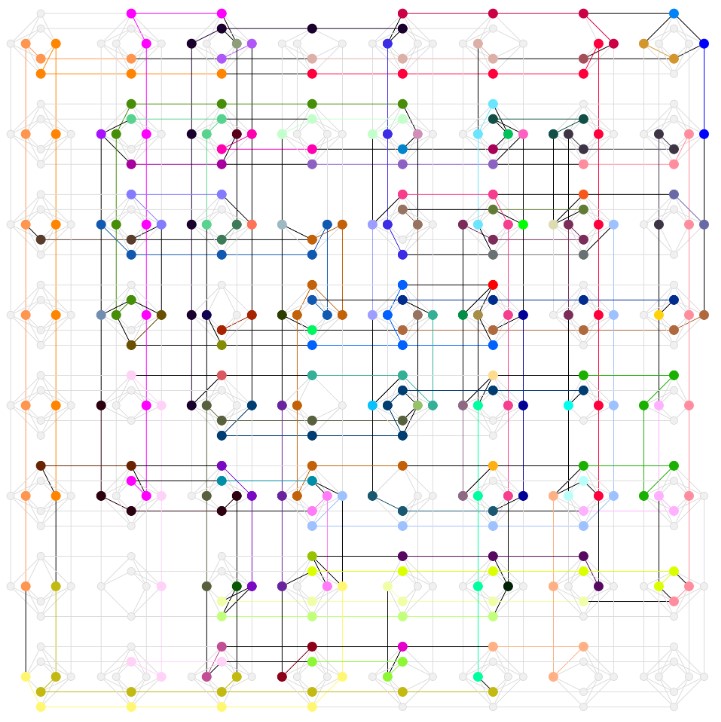}}
    \vspace{2mm}
    \caption{Graphical representation of the minor embedding step. In the graph shown in (a), each node represents a binary variable and each edge represents a logical connection between two variables of the \ac{QUBO} problem. In (b), each qubit chain is defined by the color of the embedded variable. Source: \cite{McGeoch2014}}
    \label{fig:QUBO_embedding}
\end{figure*}

\section{Background}
\label{sec:background}

\subsection{Quantum Annealing and QUBO}
\label{sub:QA}
To understand the underlying working principles of D-Wave quantum annealers, a brief introduction is needed. In \ac{AQC} \cite{McGeoch2014,Albash2018}, the forces acting in a quantum system are described by a time-varying Hamiltonian $\mathcal{H}(t)$.
The time evolution of the state of a quantum system $\ket{\varphi(t)}$ is described by the Schr\"odinger's Equation:
\begin{equation}
\label{eq:AQC}
i\hbar\frac{\partial\ket{\varphi(t)}}{\partial t} = \mathcal{H}(t)\ket{\varphi(t)}
\end{equation}
where $i$ is the immaginary unit and $\hbar$ is the reduced Planck constant.
During the adiabatic evolution, the Hamiltonian gradually transitions from the initial Hamiltonian $\mathcal{H}_I$ to the final Hamiltonian $\mathcal{H}_F$:
\begin{equation}
\label{eq:AQC-Hamiltonian}
    \mathcal{H}(t)=s(t)\mathcal{H}_I+(1-s(t))\mathcal{H}_F
\end{equation}
where $s(t)$ is a function modeling the transition, such that $s(0)=1$ and $s(t_f)=0$ after a certain elapsed time $t_f$. Given the assumptions of the adiabatic theorem \cite{Born1928}, during the time evolution, the quantum system remains at ground state, i.e., the state with lowest energy associated with the Hamiltonian. The idea in \ac{AQC} is to encode the desired result as the ground state of the final Hamiltonian $\mathcal{H}_F$.

\ac{QA} falls into the category of \ac{AQC} algorithms. More precisely, it is a heuristic approach for solving combinatorial optimization problems. In this case, the Hamiltonian of the system is defined as:
\begin{equation}
\label{eq:QA-Hamiltonian}
    \mathcal{H}(t)=\mathcal{H}_F+\Gamma(t)\mathcal{H}_D
\end{equation}
where $\mathcal{H}_F$ is the final Hamiltonian, $\Gamma(t)$ is the \textit{transverse field coefficient} as a function of time $t$, and $\mathcal{H}_D$ is the \textit{transverse field Hamiltonian} (also called \textit{disorder Hamiltonian}). $\mathcal{H}_F$ encodes the objective function and its ground state is the solution of the optimization problem. $\Gamma(t)$ is a decreasing function, equal to $0$ for $t=t_f$. It controls the contribution of $\mathcal{H}_D$, which enables traversibility of the solution space, making the optimization process escape local minima. As for this aspect, \ac{QA} presents a similarity with simulated annealing \cite{vanLaarhoven1987}, where the temperature parameter $T$ resembles the role of $\Gamma(t)$. In this framework, the assumptions of the adiabatic theorem are relaxed, i.e., there is no requirement for the quantum system to be closed and to operate in the ground state.
The implementation of \ac{QA} provided by D-Wave quantum annealers enables the solution of a specific type of optimization problems, called \acf{QUBO} problems. A \ac{QUBO} problem is defined as:
\begin{equation}
    \label{eq:energy_function}
    \text{minimize} \quad \sum_{i<j} a_i Q_{ij} a_j
\end{equation}
where $a_i\in\{0,1\}$ are the binary variables of the problem and $Q$ an upper-triangular matrix called \ac{QUBO} matrix.

\subsection{Minor Embedding}
\label{sub:embedding}
Some restrictions on the \ac{QUBO} problems submitted to D-Wave quantum annealers are enforced, which are related to the physical qubit architecture. The qubit connectivity parameter describes how many physical connections, i.e., couplers, are implemented for each qubit. D-Wave Advantage is based on the Pegasus architecture and presents approximately $5000$ qubits, $35000$ couplers and a qubit connectivity of 15 \cite{McGeoch2021}.
When the \ac{QUBO} problem is submitted to a quantum annealer, a step called \textit{minor embedding} \cite{Cai2014} is performed. 
In the minor embedding step, a qubit chain is assigned to each binary variable of the problem $a_i$. The main requirement is maintaining the logical structure of the problem, described by $Q$. Each element of the \ac{QUBO} matrix $Q_{ij}$ represents a logical relation between the variables $a_i$ and $a_j$. The coefficients $Q_{ij}$ are mapped to the strength of the couplers connecting the qubit chains assigned to the variables $a_i$ and $a_j$. The existence of such an embedding is a requirement for a problem to be solved by the annealer, i.e., constraints on the dimension and the structure of the \ac{QUBO} problem need to be satisfied. In Fig. \ref{fig:QUBO_embedding}, the embedding of a \ac{QUBO} problem in graph form is shown.

\begin{figure*}
    \centering
    \includegraphics[width=0.80\linewidth]{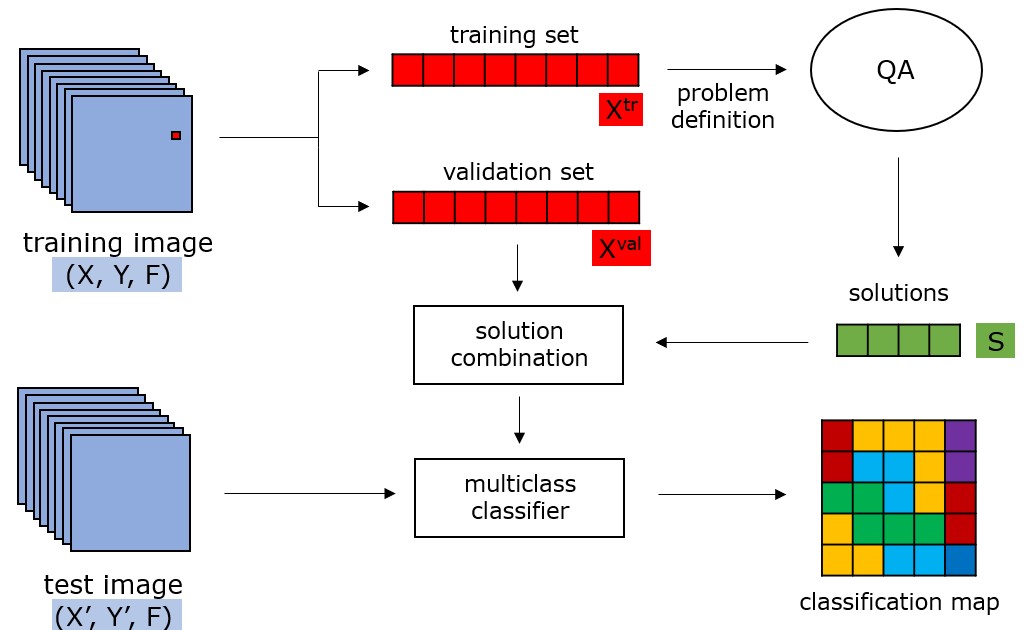}
    \caption{Workflow for the \ac{QMSVM} algorithm. A training set $X^{tr}$ is given as input to the \ac{QA} step, which obtains a set of $S$ solutions to the training problem. The solutions are then combined according to the accuracy performance on a validation set $X^{val}$, to generate the final classifier.
    }
    \label{fig:method_scheme}
\end{figure*}

\section{Quantum Multiclass SVM Formulation}
\label{sec:formulation}

In this section, a novel algorithm called \ac{QMSVM} is described. It is based on a reformulation of the \acf{CS} \ac{SVM} \cite{Crammer2002a} as a \ac{QUBO} problem. The followed steps are adapted from the \ac{QSVM} proposed in \cite{Willsch2019}, with the addition of a solution combination method. As a starting point, the \ac{CS} \ac{SVM} formulation is described in the following.

\subsection{Crammer-Singer Multiclass SVM}
\label{sub:CS-SVM}

In a supervised multiclass classification problem, let N be the number of training examples, C the number of classes, $X^{tr}=\{\mathbf{x}_n\}$ the feature vectors of dimension $F$, $Y^{tr}=\{y_n\}$ the labels. The training consists in the solution of the following quadratic program:
\begin{equation}
\begin{split}
\label{eq:multiclass_opt}
    \text{minimize} \quad F(T) = &\frac{1}{2} \sum_{n_1,n_2=0}^{N-1} K(\mathbf{x}_{n_1},\mathbf{x}_{n_2}) \sum_{c=0}^{C-1} \tau_{n_1c}\tau_{n_2c}
    \\&- \beta \sum_{n=0}^{N-1} \sum_{c=0}^{C-1} \delta_{cy_n}\tau_{nc}
\end{split}
\end{equation}
\begin{equation}
\label{eq:multiclass_constraints}
    \text{subject to} \quad
    \sum_{c=0}^{C-1} \tau_{nc} = 0 \quad \forall n, \quad
    \tau_{nc} \leq 0 \quad \forall  n, \forall c \neq y_n.
\end{equation}
where $T = [\tau_{nc}]$ is the matrix of the $NC$ problem variables, with $n=0,...,N-1$, $c=0,...,C-1$ and $\tau_{nc} \in [-1,1]$, $\delta_{ij}$ is the Kronecker delta and $\beta$ a regularization parameter.

D-Wave Advantage is unable to directly solve Eq. (\ref{eq:multiclass_opt})-(\ref{eq:multiclass_constraints}). Therefore, a reformulation of Eq. (\ref{eq:multiclass_opt})-(\ref{eq:multiclass_constraints}) as a \ac{QUBO} problem is necessary. The followed steps are: choosing a binary encoding (Sect. \ref{sub:encoding}), defining the penalty terms (Sect. \ref{sub:penalty}), deriving the QUBO matrix including the results of the previous steps in the cost function (Sect. \ref{sub:QUBO}), and defining a solution combination method (Sect. \ref{sub:solution_combination}).

\subsection{Binary Encoding}
\label{sub:encoding}
The first step consists in defining the binary variables $a_i$ of the \ac{QUBO} problem. In the \ac{CS} formulation, the problem variables $\tau_{nc}$ are real numbers. The idea is to discretize the solution space using uniform sampling and represent each value as a set of $B$ binary variables. First, the following intermediate variable is defined:
\begin{equation}
\label{eq:binary_encoding_sigma}
    \sigma_{nc} = \sum_{b=0}^{B-1} 2^b a_{nCB+cB+b}.
\end{equation}
$\sigma_{nc}$ is an integer value in $[0,2^B-1]$ represented by the binary encoding $\{a_{nCB+cB+b}\},b=0,\dots,B-1$. Then, the problem variables $\tau_{nc}$ can be defined from $\sigma_{nc}$ as:
\begin{equation}
\label{eq:binary_encoding_tau}
    \tau_{nc} = -1 + \frac{2}{2^{B}-1} \sigma_{nc} =-1+\frac{2}{2^{B}-1} \sum_{b=0}^{B-1} 2^b a_{nCB+cB+b}.
\end{equation}
With this definition, it can be proven that each value of $\tau_{nc}$ lies in $[-1,1]$ and the interval is uniformly sampled.

Fig. \ref{fig:sampling} shows the sampling of Eq. (\ref{eq:binary_encoding_tau}) for $B=2$, i.e., in the case each sample of $\tau_{nc}$ is represented by 2 binary variables, indicated above each sample. Since the total number of problem variables is $NC$ (each variable is associated with an example and a class), the whole optimization space can be represented by a set of $NCB$ binary variables $\{a_0,a_1,...,a_{NCB-1}\}$.
\vspace{0.5cm}
\begin{figure}
\begin{center}
\begin{tikzpicture}[scale=0.75]
    \draw[->] (-0.75,0) -- (6.75,0) node [right] {$\sigma_{nc}$};

    \foreach \pos/\label in {0/$0$, 2/$1$,
            4/$2$, 6/$3$}
        \draw (\pos,0) -- (\pos,-0.3) (\pos cm,-5ex) node
            [anchor=base,fill=white,inner sep=1pt]  {\label};
    \draw (0,0) node[circle,fill,inner sep=2pt,label=above:\{00\}]{};
    \draw (2,0) node[circle,fill,inner sep=2pt,label=above:\{01\}]{};
    \draw (4,0) node[circle,fill,inner sep=2pt,label=above:\{10\}]{};
    \draw (6,0) node[circle,fill,inner sep=2pt,label=above:\{11\}]{};

\end{tikzpicture}
\qquad
\begin{tikzpicture}[scale=0.75]
    \draw[->] (-1,0) -- (5,0) node [right] {$\tau_{nc}$};

    \foreach \pos/\label in {0/$-1$, 2/$0$,
            4/$1$}
        \draw (\pos,0) -- (\pos,-0.3) (\pos cm,-5ex) node
            [anchor=base,fill=white,inner sep=1pt]  {\label};
    \draw (0,0) node[circle,fill,inner sep=2pt,label=above:\{00\}]{};
    \draw (1.33,0) node[circle,fill,inner sep=2pt,label=above:\{01\}]{};
    \draw (2.67,0) node[circle,fill,inner sep=2pt,label=above:\{10\}]{};
    \draw (4,0) node[circle,fill,inner sep=2pt,label=above:\{11\}]{};
\end{tikzpicture}
\caption{Representation of the chosen variable sampling and encoding for $B=2$.}
\label{fig:sampling}
\end{center}
\end{figure}
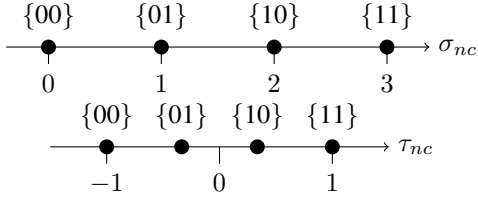
\subsection{Penalty Terms}
\label{sub:penalty}
Another requirement is to include the constraints of Eq. (\ref{eq:multiclass_constraints}), as no constraints can be directly enforced in a \ac{QUBO} problem. A possibility is to add the constraints to the \ac{QUBO} matrix as weighted positive penalty terms. For the first constraint, the penalty term needs to increase in the case the difference between the value of the sum and $0$ increases. In addition, a penalty term needs to be associated with each training example and with the same weight. Since a quadratic polynomial term is required, the following penalty term is chosen:
\begin{equation}
\label{eq:multiclass_constraints_qubo_1}
    P_n^1=\left( \sum_{c=0}^{C-1} \tau_{n,c} \right)^2
\end{equation}
For the second constraint, which is an inequality, it is sufficient to directly consider $\tau_{nc}$ as the penalty term associated with each training sample and each class. A coefficient $(1-\delta_{cy_n})$ is attached to account for the case $c=y_n$, in which the penalty is zero:
\begin{equation}
\label{eq:multiclass_constraints_qubo_2}
    P_{nc}^2=(1-\delta_{cy_n})\tau_{n,c}
\end{equation}
The final penalty term can be written as:
\begin{equation}
\label{eq:multiclass_constraints_qubo}
\begin{split}
    P&=\sum_{n=0}^{N-1} P_n^1 + \sum_{n=0}^{N-1} \sum_{c=0}^{C-1} P_{nc}^2
    \\&=\sum_{n=0}^{N-1} \left( \sum_{c=0}^{C-1} \tau_{n,c} \right)^2
    + \sum_{n=0}^{N-1} \left( \sum_{c=0}^{C-1} (1-\delta_{cy_n})\tau_{n,c} \right)
\end{split}
\end{equation}
Note that $P_n^1$ and $P_{nc}^2$ are included with the same weight. The following reasons behind this choice can be listed:
\begin{itemize}
    \item Considering two different weights would increase the number of hyperparameters of the optimization problem and the already high complexity of the tuning phase;
    \item The two constraints have to be both equally satisfied;
    \item The two penalty terms have approximately the same order of magnitude, as $\tau_{nc}\in[-1,1]$, so there is no imbalance in values.
\end{itemize}

\subsection{QUBO Matrix}
\label{sub:QUBO}
The QUBO problem can be now written by adding to Eq. (\ref{eq:multiclass_opt}) the penalty term in Eq.  (\ref{eq:multiclass_constraints_qubo}) multiplied by a weight $\mu$ and substituting $\tau_{i,j}$ with the encoding in Eq. (\ref{eq:binary_encoding_tau}). The energy function $E$ can be written in the following form\footnote{In this formulation, a simplified notation for the sums is used, as the range of the indices is unaltered and redundant.}:
\begin{equation}
\begin{split}
\label{eq:msvm_energy_function}
    E= \enskip &F+\mu P
    = \enskip \sum_{n_1n_2c_1c_2b_1b_2} a_{n_1CB+c_1B+b_1}\\ &\widetilde{Q}_{n_1CB+c_1B+b_1,n_2CB+c_2B+b_2} a_{n_2CB+c_2B+b_2}.
\end{split}
\end{equation}
$\widetilde{Q}$ is a symmetric matrix of size $NCB\times NCB$. It can be analytically derived by neglecting the terms not depending on the binary variables and is equal to:
\begin{equation}
\begin{split}
\label{eq:msvm_qubo_matrix}
    &\widetilde{Q}_{n_1CB+c_1B+b_1,n_2CB+c_2B+b_2} =
    \\= \enskip &\delta_{n_1n_2}\delta_{c_1c_2}\delta_{b_1b_2} \frac{2^{b_1+1}}{2^B-1}\Aperta -\sum_{i}K(\mathbf{x}_{n_1},\mathbf{x}_{i})
    \\&- \delta_{c_1y_{n_1}}  \left(\beta+\mu\right)- 2C\mu + \mu \Chiusa
    \\&+ \delta_{c_1c_2}\frac{2^{b_1+b_2+1}}{(2^B-1)^2} K(\mathbf{x}_{n_1},\mathbf{x}_{n_2}) + \delta_{n_1n_2}\frac{2^{b_1+b_2+2}\mu}{(2^B-1)^2}
\end{split}
\end{equation}

The upper-triangular QUBO matrix $Q$ can be computed from $\widetilde{Q}$ as:
\begin{equation}
\label{eq:msvm_upper_triangular}
    Q_{i j}=
    \begin{cases}
      \widetilde{Q}_{i j} & \text{for}\ i=j\\
      \widetilde{Q}_{i j}+\widetilde{Q}_{j i} & \text{for}\ i<j\\
      0 & \text{otherwise}
    \end{cases}
\end{equation}

\subsection{Solution Combination}
\label{sub:solution_combination}

Once the QUBO matrix is defined, the problem can submitted to the quantum annealer, assuming the existence of an embedding. As the annealing process is performed multiple times, depending on the value of \textit{num\_reads}, the obtained output is a set of \textit{num\_reads} solutions. The best $S$ solutions are selected, i.e., $T_i = [\tau_{nc}]_i, i=0,\dots,S-1$, ranked by the value of the energy function $E(T_i)$.
During the experiments, it has been noticed that there is no perfect correlation between solutions with lower energy and better classification accuracy of the obtained classifier. Note also that the solution space investigated by the quantum annealing algorithm is discrete, due to the variable sampling, so the obtained individual solutions can be sub-optimal. For these reasons, a solution combination is performed in order to obtain an optimal final solution. A weighted average is performed, where the weights $w_s$ for each solution $s$ are set according to the prediction accuracy of the obtained classifiers on a validation set $\{X^{val},Y^{val}\}$. In particular, the solutions above a certain threshold accuracy are selected, and their weight is computed applying the softmax function to $\textit{multiplier} \cdot \textit{accuracy}_s$, where $\textit{multiplier}$ is a real value and $\textit{accuracy}_s$ is the accuracy of the classifier defined by the $s$-th solution on the whole training set. The rest of the weights are set to $0$. The combined solution is computed as:
\begin{equation}
\label{eq:solution_combination}
    \bar{T}=\frac{1}{S}\sum_{s=0}^{S-1} w_s T_s.
\end{equation}
The resulting variables $\bar{\tau}_{nc}$ are then used to classify new examples:
\begin{equation}
\label{eq:classifier}
    H(\mathbf{x})= \arg \max_{c} \Bigg\{\sum_{n=0}^{N-1} \bar{\tau}_{nc} K(\mathbf{x}, \mathbf{x}_n) \Bigg\}
\end{equation}
Alg. \ref{alg:QMSVM} summarizes the implemented computational steps required for the training.

\begin{algorithm}
\caption{Quantum Multiclass SVM (QMSVM)}
\begin{algorithmic}[1]
\renewcommand{\algorithmicrequire}{\textbf{Input:}}
\renewcommand{\algorithmicensure}{\textbf{Output:}}
\REQUIRE $X^{tr}, Y^{tr}, X^{val}, Y^{val}, C, B, \beta, \mu, \gamma, S, \text{multiplier}$
\ENSURE  $\bar{T}$
\\ \textit{QUBO matrix initialization}, eq. (\ref{eq:msvm_qubo_matrix})-(\ref{eq:msvm_upper_triangular})
\STATE $Q \gets \text{QUBO\_MATRIX}(X^{tr}, Y^{tr}, C, B, \beta, \mu, \gamma)$
\\ \textit{Run annealing step and sample \textup{num\_reads} solutions}
\STATE $T \gets \text{QUANTUM\_ANNEALING}(Q)$
\\ \textit{Evaluate classifier on validation set}, eq. (\ref{eq:classifier})
\FOR {$s = 1$ to $S$}
\STATE $\hat{Y}^{val}[s] \gets H(X^{val}, Y^{val}, T)$
\STATE $\text{accuracy}[s] \gets \text{ACCURACY}(\hat{Y}^{val}[s],Y^{val})$
\ENDFOR
\\ \textit{Weights calculation}
\STATE $\text{threshold} \gets \text{THRESHOLD}(\text{accuracy})$
\FOR {$s = 1$ to $S$}
\IF {$\text{accuracy}[s]<\text{threshold}$}
\STATE $\text{accuracy}[s] \gets 0$
\ENDIF
\ENDFOR
\STATE $W \gets \text{SOFTMAX}(\text{multiplier} \cdot \text{accuracy})$
\\ \textit{Solution combination}, eq. (\ref{eq:solution_combination})
\STATE $\bar{T}\gets\text{COMBINE}(T,W)$
\RETURN $\bar{T}$
\end{algorithmic}
\label{alg:QMSVM}
\end{algorithm}

\begin{table*}[!t]
\caption{Datasets Used in the Experiments.}
\label{tab:datasets}
\vspace{2mm}
\centering
\renewcommand{\arraystretch}{1.3}
\begin{tabular}{*{7}{c}}
    \hline
    \textbf{Dataset} & \textbf{Dimension} & \textbf{Resolution} & \textbf{Features} & \textbf{Classes} & \textbf{Training Set} & \textbf{Test Set} \\
    \hline
    SemCity Toulouse \cite{Roscher2020} & $16$ tiles, $3504\times3452$ & $2$ m & $8$ bands & $7$ & $N$ samples, tile $4$ & $800\times800$ area, tile $8$\\
    ISPRS Potsdam \cite{Potsdam} & $38$ tiles, $6000\times6000$ & $5$ cm & $4$ bands + DSM & $5$ & $N$ samples, Tile $6\_9$ & $1000\times1000$ area, tile $6\_10$\\
    \hline
\end{tabular}
\end{table*}

\begin{table*}[!t]
\caption{Parameters Setup.}
\label{tab:parameters}
\vspace{2mm}
\centering
\renewcommand{\arraystretch}{1.3}
\begin{tabular}{*{3}{c}}
    \hline
    \textbf{Variable Name} & \textbf{Value} & \textbf{Description} \\
    \hline
    $C$ & $3$ & Number of classes considered in the multiclass classification problem \\
    $B$ & $2$ & Number of binary variables representing each problem variable $\tau_{nc}$ \\
    $\beta$ & $1$ & Model related regularization parameter, introduced in Eq. (\ref{eq:multiclass_opt}) \\
    $\mu$ & $1$ & Weight of the penalty term $P$ added to the energy function defined in Eq. (\ref{eq:msvm_energy_function}) \\
    $\gamma$ & $1$ & Gaussian kernel parameter, regulating its radius \\
    $N$ & $[50,40000]$ & Total number of training examples used in the training (multiple values analyzed) \\
    $M$ & $60$ & Number of selected training examples, used to define the \ac{QUBO} matrix $Q$ submitted to the \ac{QA} \\
    \textit{num\_reads} & $1000$ & Number of times the annealing schedule is performed and how many solutions are sampled for each run \\
    $S$ & $100$ & Number of solutions selected among the total number \textit{num\_reads}, used for the solution combination \\
    \textit{multiplier} & $10$ & Regulates the balance between higher and lower accuracy values over the combined solutions \\
    \textit{max\_min\_ratio} & $15$ & Ratio between the maximum and minimum non-zero absolute value of the \ac{QUBO} matrix $Q$ \\
    \textit{chain\_strength} & $1$ & Relative coupling strength between qubits that form a chain and qubits in different chains \\
    \textit{annealing\_time} & $200$ & Time (in \textmu s) at which the measurement is performed after starting the annealing schedule \\
    \hline
\end{tabular}
\end{table*}

\section{Evaluation}
\label{sec:evaluation}
\subsection{Experimental Setup}
\label{sub:setup}

\begin{figure}
    \centering
    \includegraphics[width=\linewidth]{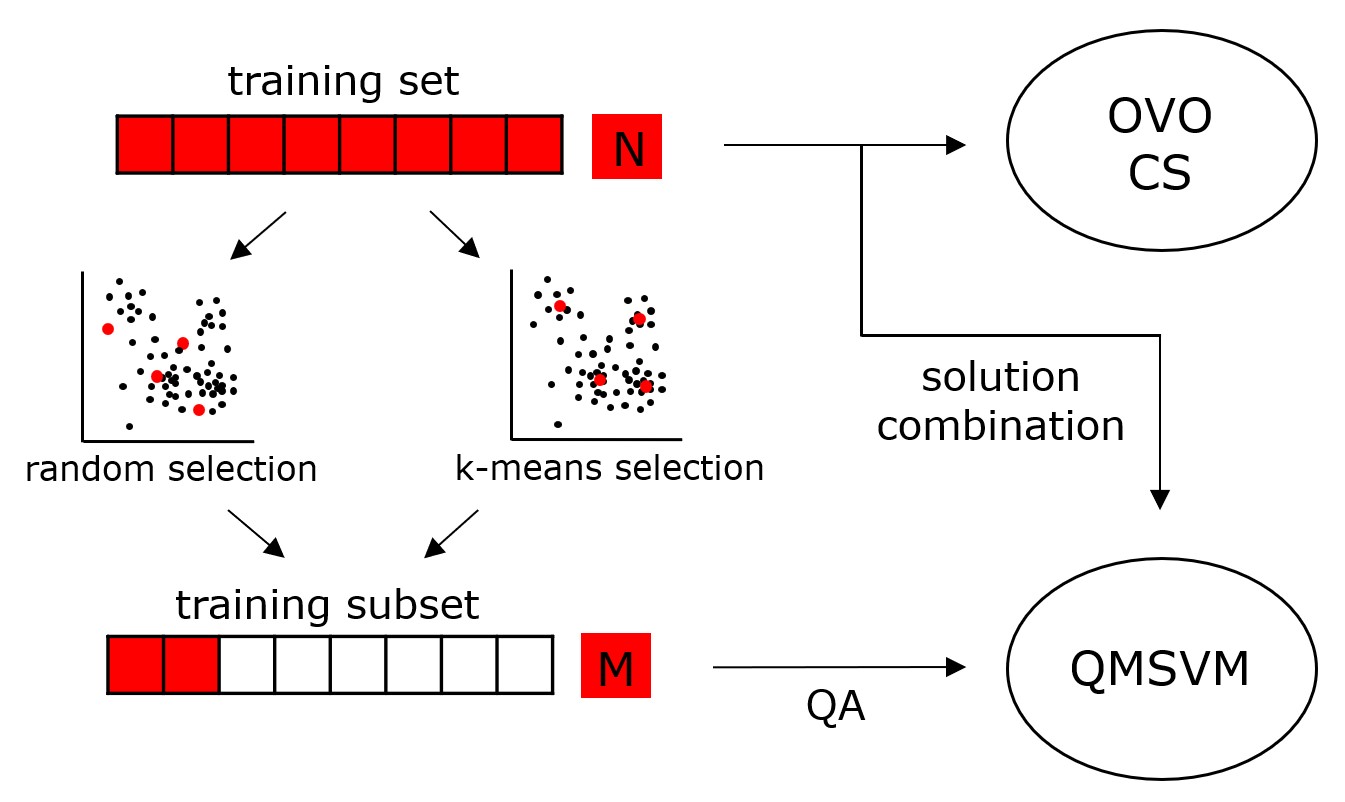}
    \caption{Training setup. The \ac{OVO} and \ac{CS} methods have been trained on a training set of $N$ examples. For \ac{QMSVM}, a subset of $M$ examples is selected and used by the annealing algorithm, while the solution combination is performed based on the accuracy obtained on the whole training set.
    }
    \label{fig:example_selection}
\end{figure}

The QMSVM algorithm has been validated on a semantic segmentation problem applied to multispectral \ac{RS} images. Two different datasets are considered, i.e., SemCity Toulouse \cite{Roscher2020} (hereafter ``Toulouse") and ISPRS Potsdam \cite{Potsdam} (hereafter ``Potsdam"). Tab. \ref{tab:datasets} describes the selected datasets. While both represent urban areas, the two datasets differ in the used features and the ground resolution. From each dataset, a training set of $N$ examples is initialized.

The experiments have been performed on a real quantum device, JUPSI \cite{JUPSI}, a D-Wave Advantage quantum annealer located at Forschungszentrum J\"ulich. The \textit{Advantage\_system5.3} cloud solver has been used.
Given the memory and connectivity limitation of the machine, the training set $\{X^{tr},Y^{tr}\}$ defined in Sect. \ref{sub:CS-SVM} is initialized as a subset of $M$ examples from the total number of training examples $N$. 
The training subset is computed through an example selection step, and two selection methods have been tested.
\begin{itemize}
    \item{\textit{Random selection}: $M$ random examples are selected from the whole training set. It is a fast and straightforward method, enforcing no selection criterion.}
    \item{\textit{K-means selection}: k-means clustering \cite{Hartigan1979} is applied to each of the $C$ classes, with $k=\frac{M}{C}$, and the obtained $M$ centroids are used as selected examples. It is inspired by undersampling techniques in imbalanced classification \cite{Huang2006}. In principle, the method is designed to select meaningful examples, covering the whole feature domain.}
\end{itemize}
The whole training set is then used as the validation set $\{X^{val},Y^{val}\}$ for the solution combination, introduced in Sect. \ref{sub:solution_combination}. The \textit{threshold} accuracy, which determines which solutions are discarded in the combination, has been computed as:
\begin{equation}
    \text{threshold}=0.2\cdot\min(\text{accuracy})+0.8\cdot\max(\text{accuracy}).
\end{equation}

The results are compared with two standard implementations of the \ac{MSVM}, i.e., the \ac{OVO} implementation in Scikit-learn \cite{Pedregosa2011} and a \ac{CS} \ac{SVM} implementation in C++ \cite{MSVM}. The training setup is depicted in Fig. \ref{fig:example_selection}.

\renewcommand\figurespace{0.3\linewidth}
\renewcommand\figurewidth{0.22\linewidth}
\begin{figure*}
    \centering
    \subfloat[Ground truth]{\fbox{\includegraphics[width=\figurewidth]{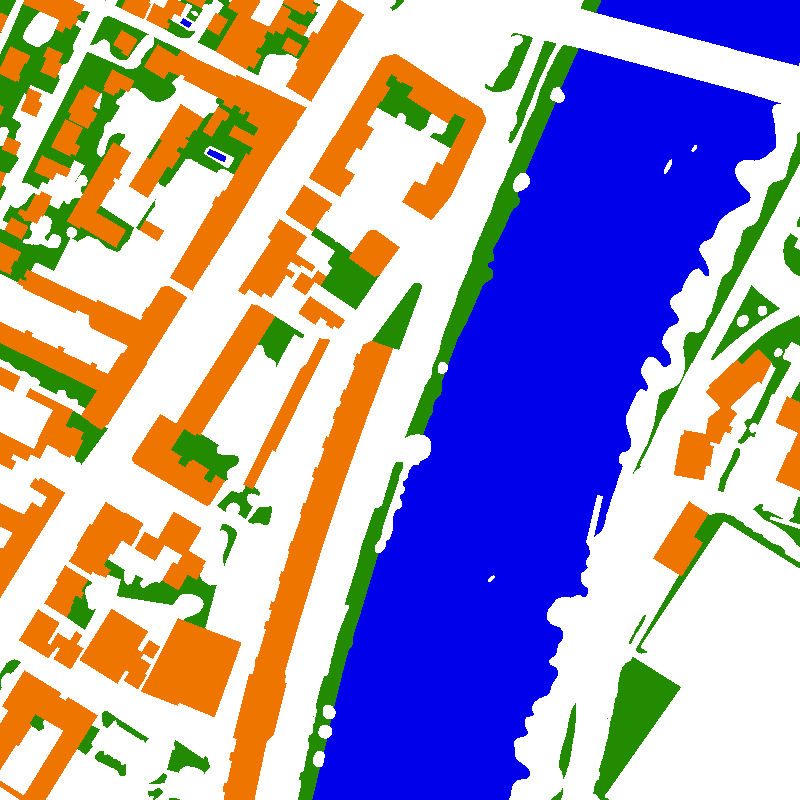}}%
    }
    \hfil
    \subfloat[OVO]{\fbox{\includegraphics[width=\figurewidth]{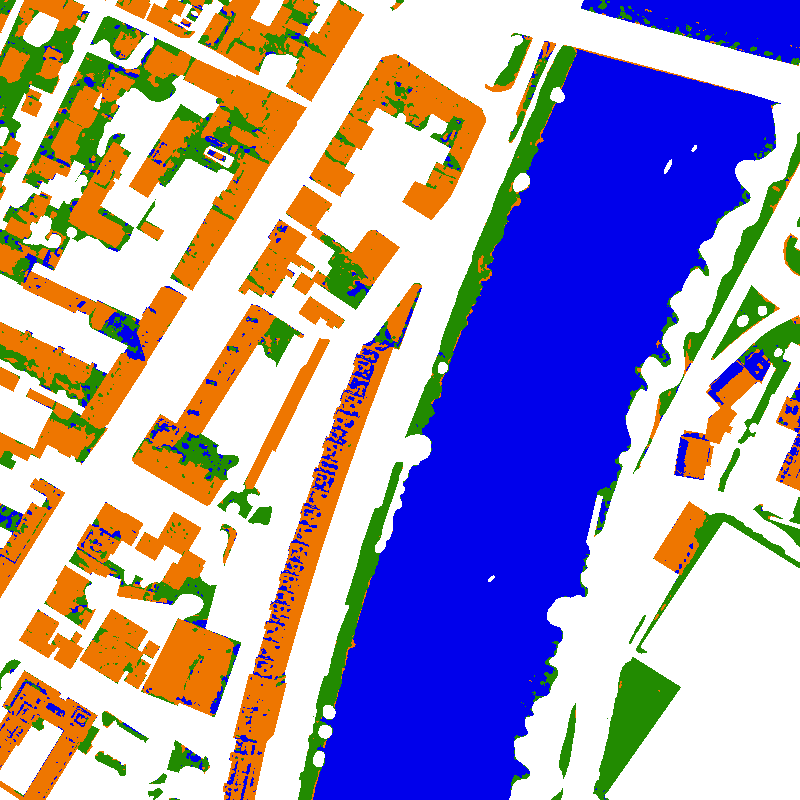}}%
    }
    \hfil
    \subfloat[CS]{\fbox{\includegraphics[width=\figurewidth]{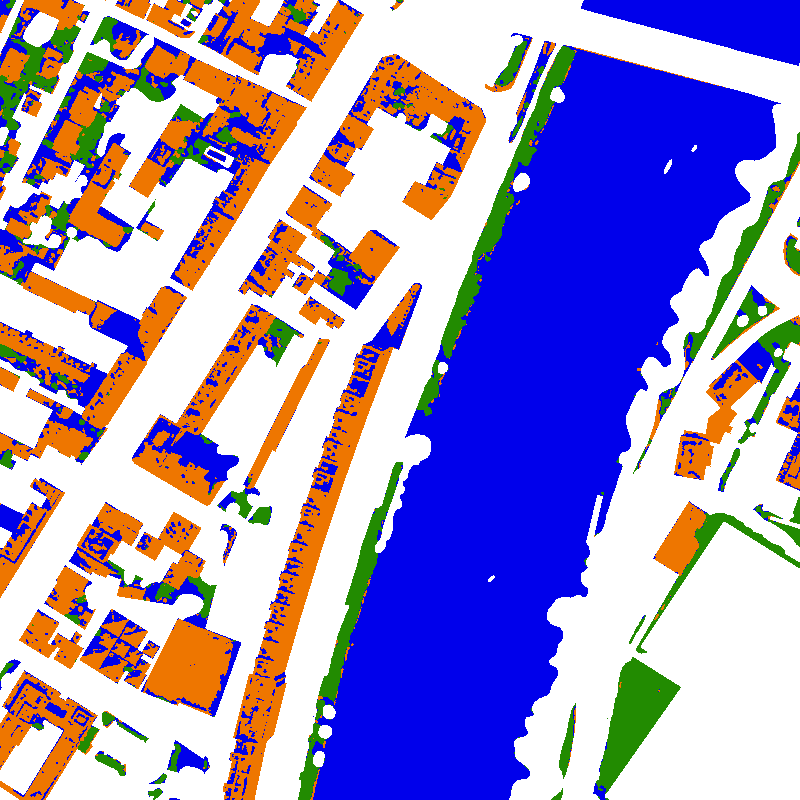}}%
    }
    \hfil
    \subfloat[QMSVM]{\fbox{\includegraphics[width=\figurewidth]{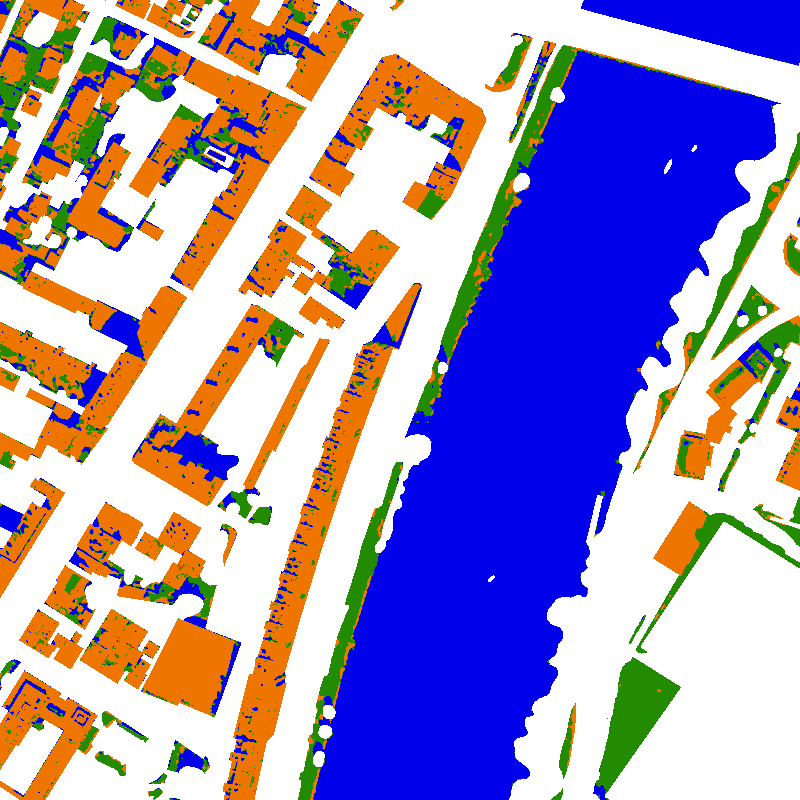}}%
    }
    \vspace{2mm}
    \caption{Toulouse - ground truth and predicted land cover maps on an $800\times800$ selected area from tile 8 for \acf{OVO}, \acf{CS} and \acf{QMSVM} using $N=40000$ training examples. The considered classes are "building" (orange), "pervious surface" (green) and "water" (blue).}
    \label{fig:landcover_toulouse}
\end{figure*}
\renewcommand\figurespace{0.3\linewidth}
\renewcommand\figurewidth{0.22\linewidth}
\begin{figure*}
    \centering
    \subfloat[Ground truth]{\fbox{\includegraphics[width=\figurewidth]{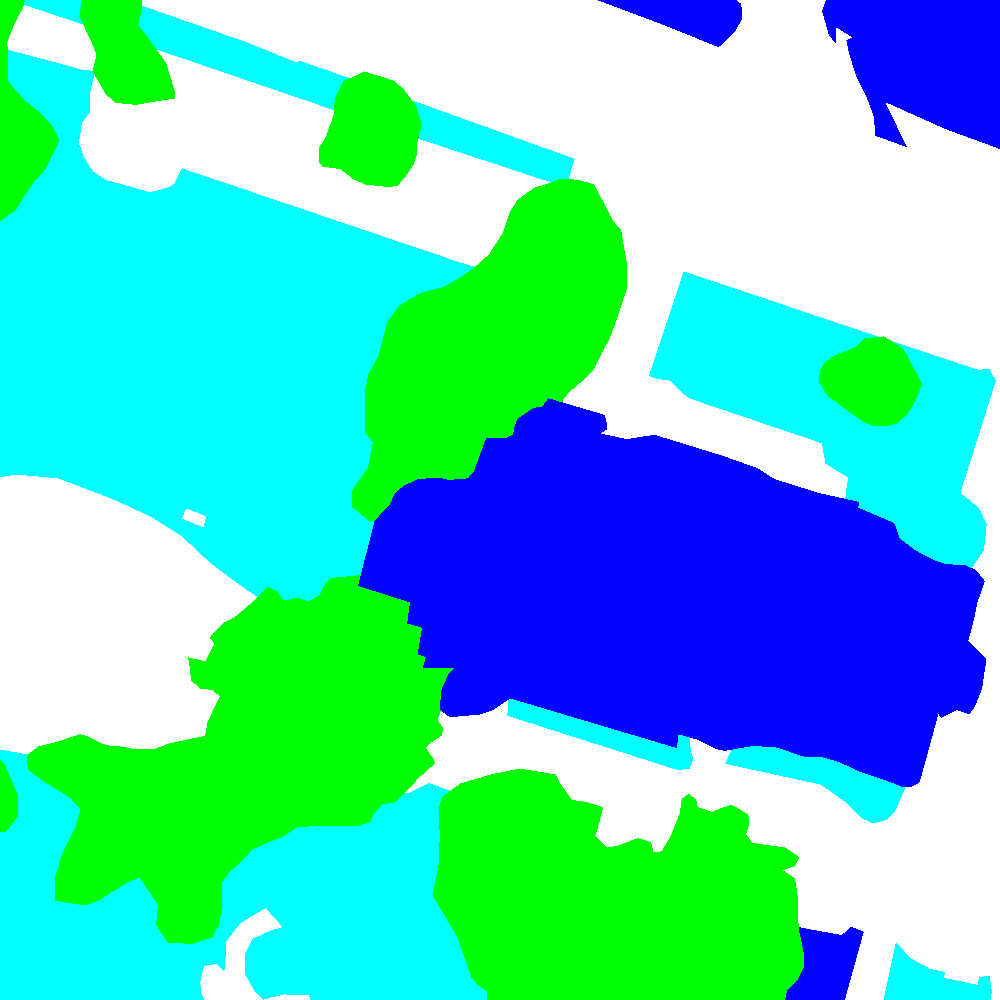}}%
    }
    \hfil
    \subfloat[OVO]{\fbox{\includegraphics[width=\figurewidth]{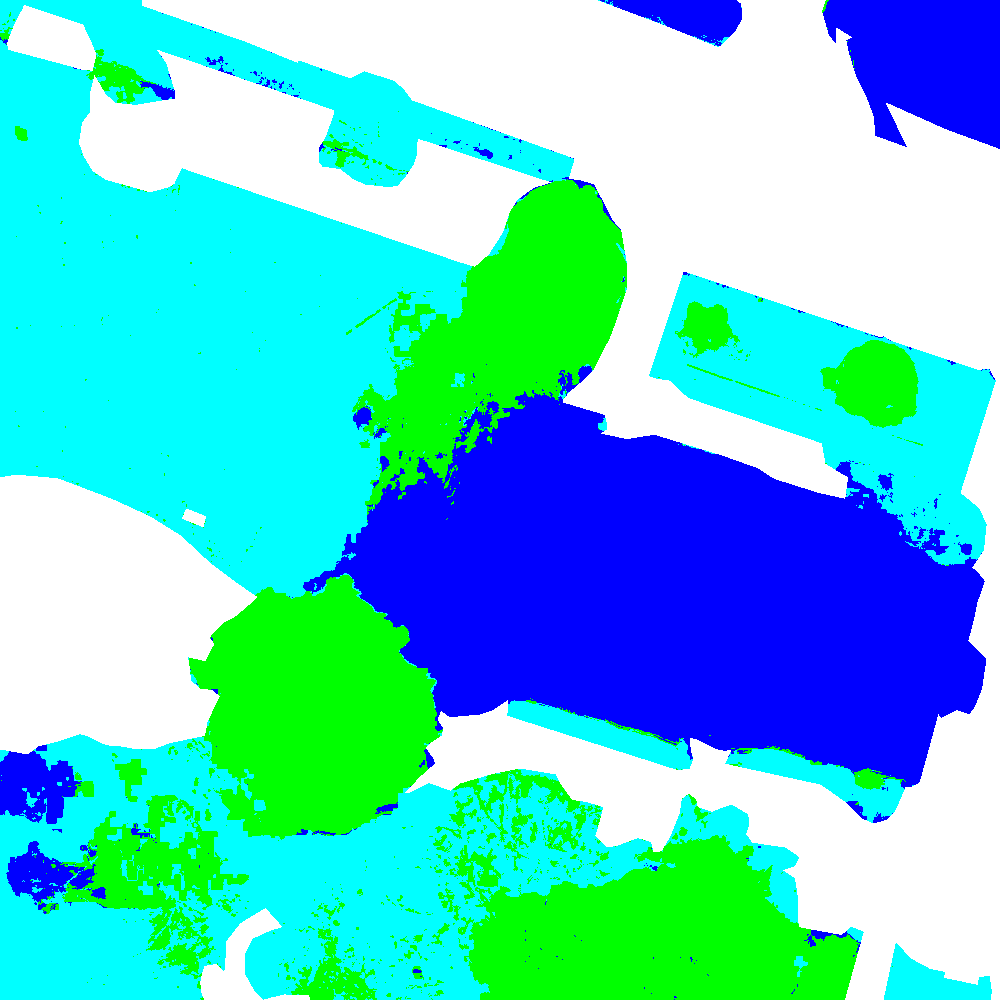}}%
    }
    \hfil
    \subfloat[CS]{\fbox{\includegraphics[width=\figurewidth]{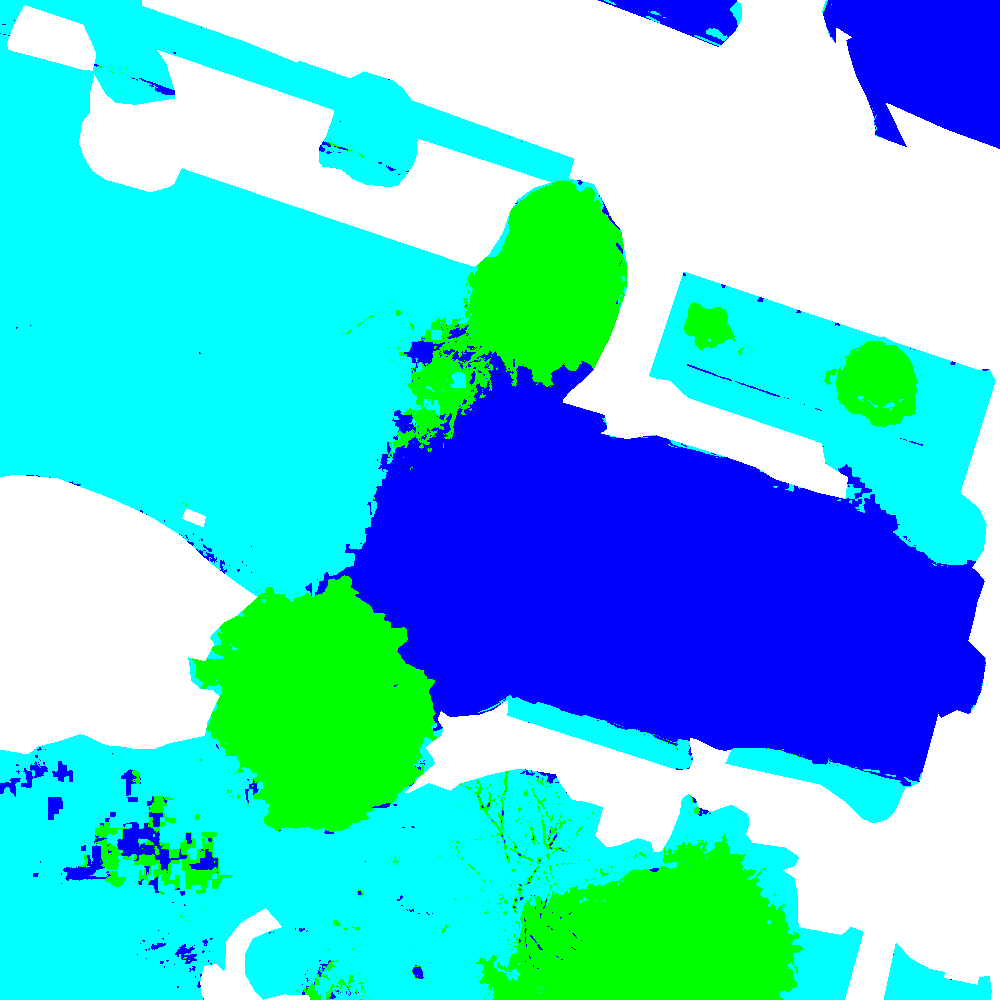}}%
    }
    \hfil
    \subfloat[QMSVM]{\fbox{\includegraphics[width=\figurewidth]{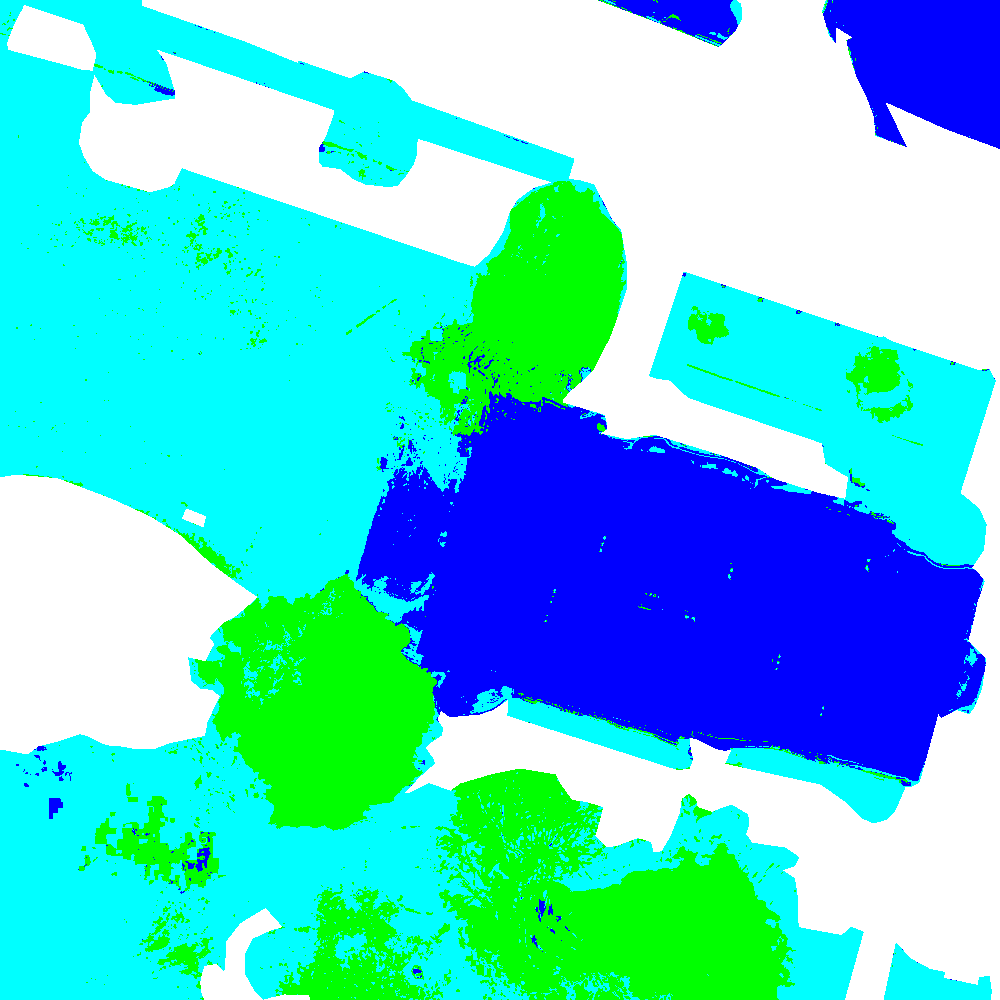}}%
    }
    \vspace{2mm}
    \caption{Potsdam - ground truth and predicted land cover maps on a $1000\times1000$ selected area from tile 6\_10 for \acf{OVO}, \acf{CS} and \acf{QMSVM} using $N=15000$ training examples. The considered classes are "building" (blue), "low vegetation" (light blue) and "tree" (green).}
    \label{fig:landcover_potsdam}
\end{figure*}
%
\begin{table}[!t]
\caption{Results of Accuracy, F1 Score and Execution Time for OVO and QMSVM.} 
\label{tab:results}
\vspace{2mm}
\centering
\renewcommand{\arraystretch}{1.3}
\begin{tabular}{*{7}{c}}
    \hline
    \textbf{Solver} & \textbf{N} & \textbf{M} & \textbf{Accuracy} & \textbf{F1} & \textbf{t (s)} \\
    \hline
    \multicolumn{6}{c}{\textit{Toulouse - best accuracy, maximum N}}\\
    OVO & $40000$ & - & $0.9123$ & $0.9147$ & $726.49$\\
    CS & $40000$ & - & $0.8277$ & $0.8463$ & $2951.18$\\
    QMSVM (random) & $40000$ & $60$ & $0.8803$ & $0.8881$ & $125.76$\\
    QMSVM (k-means) & $40000$ & $60$ & $0.8406$ & $0.8483$ & $168.36$\\
    \hline
    \multicolumn{6}{c}{\textit{Potsdam - best accuracy, maximum N}}\\
    OVO & $15000$ & - & $0.8556$ & $0.8599$ & $808.37$\\
    CS & $15000$ & - & $0.8226$ & $0.8362$ & $2708.62$\\
    QMSVM (random) & $15000$ & $60$ & $0.7814$ & $0.7880$ & $95.72$\\
    QMSVM (k-means) & $15000$ & $60$ & $0.7648$ & $0.7640$ & $74.36$\\
    \hline
\end{tabular}
\end{table}

\subsection{Parameters}
\label{sub:parameters}
In Tab. \ref{tab:parameters} the parameters of the problem are described.
The parameters $\beta$, $\mu$ and $\gamma$ are set through a simple grid search optimization on a validation set. Different values of $N$ are chosen in order to analyze the performance of the method by varying the number of available examples. The highest tested value is $N=40000$ for Toulouse and $N=15000$ for Potsdam.
The parameters $B$, $M$ and \textit{max\_min\_ratio} are related to the main limitation of the \ac{QA}, i.e., the number of qubits and couplers. As previously discussed, the possibility of finding an embedding on the given qubit architecture is required for solving the \ac{QUBO} problem. This is achieved in case $Q$ is sufficiently small, sufficiently sparse, or both. Using only the selected training subset, the dimension of $Q$ is $MCB$. Thus, $M$ is limited, which is the reason the \ac{QA} is unable to use an arbitrarily large training set and the example selection is performed in the first place. To maximize the number of examples fitting in the \ac{QA}, the remaining parameters are kept low, i.e., $C=3$ and $B=2$. Considering a higher number of classes, i.e., $C>3$, would require using a lower number of examples $M$, which degrades the overall performance. Regarding sparsity, a straightforward operation is performed, i.e., pruning the values of $Q$ below the threshold defined by \textit{max\_min\_ratio}. This simplification is acceptable, as relatively low values would be mapped to relatively low strengths in the \ac{QA}, which barely affect the annealing process.
The parameters \textit{num\_reads}, \textit{chain\_strength} and \textit{annealing\_time} are related to the annealer setup. Their values are chosen considering the impact they have on the quality of the obtained solutions \cite{Willsch2022}. The remaining parameters are set arbitrarily. Tab. \ref{tab:parameters} summarizes the chosen parameter values.

\subsection{Results}
\label{sub:results}

In the test phase, the methods are evaluated on a $3$-class classification problem on a subtile. For Toulouse, a $800\times800$ test subtile from tile 8 has been selected and the classes "building", "pervious surface", "water" have been considered. For Potsdam, a $1000\times1000$ test subtile from tile $6\_10$ has been selected and the classes "building", "low vegetation", "tree" have been considered.

The method is evaluated according to both test accuracy and execution time. For \ac{OVO} and \ac{CS}, training and inference are considered. For \ac{QMSVM}, the time measurement includes preprocessing (sample selection), training (annealing), postprocessing (solution combination) and inference time. 
Fig. \ref{fig:landcover_toulouse}-\ref{fig:landcover_potsdam} shows the ground truth of the selected subtile and the ground maps obtained by \ac{OVO}, \ac{CS} and \ac{QMSVM}. In Fig. \ref{fig:toulouse_accuracy_time}-\ref{fig:potsdam_alltimes} the performance of the analyzed methods on the Toulouse and Potsdam dataset is shown in terms of both test accuracy and execution time, and for both example selection methods, i.e., random and k-means selection. Tab. \ref{tab:results} summarizes the best obtained results in terms of test accuracy, along with the respective $F_1$ score. Accuracy and $F_1$ score are computed as:
\begin{equation}
    \text{accuracy} = \frac{\text{correct predictions}}{\text{no. of predictions}}, \quad
    F_1 = \text{average}_c(F_{1c}).
\end{equation}
$F_{1c}$ is the $F_1$ score computed for each class $c$ with respect to the remaining classes:
\begin{equation}
    F_{1c} = \frac{\text{TP}_c}{\text{TP}_c + \frac{1}{2}(\text{FN}_c+\text{FP}_c) },
\end{equation}
where TP (true positive), FN (false negative) and FP (false positive) predictions are referred to the class $c$.
Given the result variability for \ac{QMSVM}, related to the intrinsic randomness of the example selection, 5 runs for each selection method have been performed, each one with a different chosen random seed. For each selection method, the average result is plotted and the space between the best and the worst obtained result is highlighted.

\subsection{Analysis}
\label{sub:analysis}

From a first impression, it can be noticed that \ac{QMSVM} is a feasible quantum implementation of the \ac{CS} \ac{SVM} method. \ac{QMSVM} is able to reach a slightly lower or comparable classification accuracy with respect to \ac{CS}, despite the low number of training examples $M$ that can handle in the optimization step. The reason is that the solution combination using $N$ examples is able to improve the quality of the final solution on average by increasing $N$, as seen in the accuracy plots. However, the prediction accuracy of \ac{OVO} is slighly higher than both \ac{CS} and \ac{QMSVM} for higher $N$. A worse performance of \ac{QMSVM} can be clearly seen on the more complex Potsdam dataset, where the result variability is also higher with respect to Toulouse. In both cases, the random selection method consistently outperforms k-means selection.

Regarding the execution time, \ac{QMSVM} can better handle a high number of training examples $N$ with respect to \ac{OVO} and \ac{CS}.
The different steps included in the measurement of the execution time are shown in detail in Fig. \ref{fig:toulouse_alltimes}-\ref{fig:potsdam_alltimes}. It can be noticed that the most demanding step in the \ac{QMSVM} is the annealing. The main reason is the high time complexity of the minor embedding algorithm. However, the interesting results are the linear time increase for the solution combination and the constant time for inference with respect to $N$. The obtained results are as expected: the solutions combination requires the computation of $SCMN$ kernels, i.e., it is linear with respect to $N$, whereas the classifier used in the inference depends on the $MC$ problem variables, i.e., it is independent from $N$. The outcomes for \ac{OVO} and \ac{CS} are also coherent with \ac{SVM} theory: the training complexity of the Scikit-learn \ac{SVM} implementation is $O(N^3)$, whereas the inference time is linearly dependent on $N$ \cite{Ns2015}.

These results clearly show that \ac{QMSVM} is a much more scalable algorithm with respect to the considered training examples $N$ compared to standard \ac{MSVM} methods. Given that the annealing step, which is a fixed step unrelated to $N$, has the largest impact in terms of time, the overall execution time can be regarded as near constant. 

\renewcommand\figurespace{0.5\linewidth}
\renewcommand\figurewidth{0.4\linewidth}
\begin{figure*}
    \centering
    \subfloat[Test accuracy vs. training set size N] {\includegraphics[width=\figurewidth]{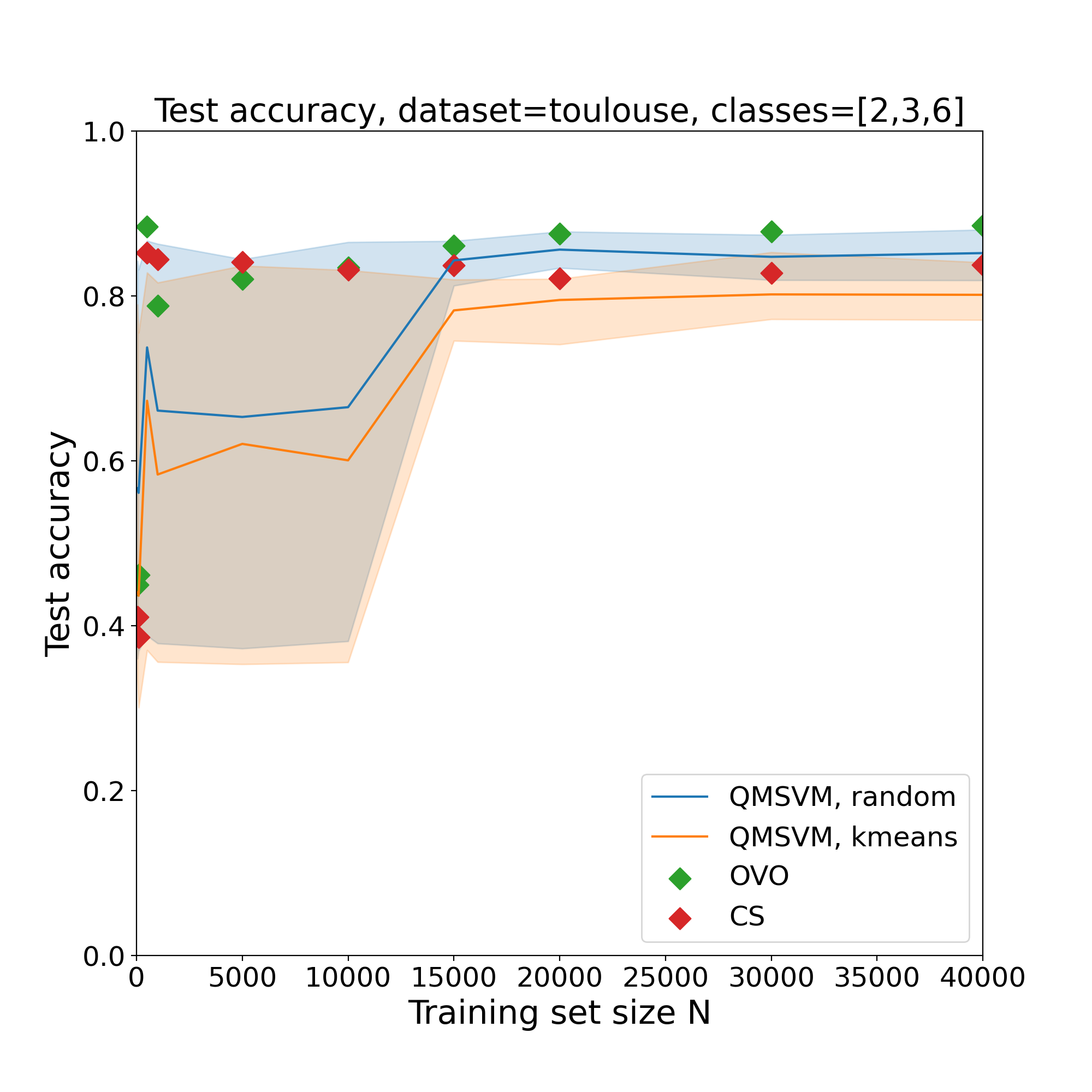}}
    \hfil
    \subfloat[Execution time vs. training set size N]{\includegraphics[width=\figurewidth]{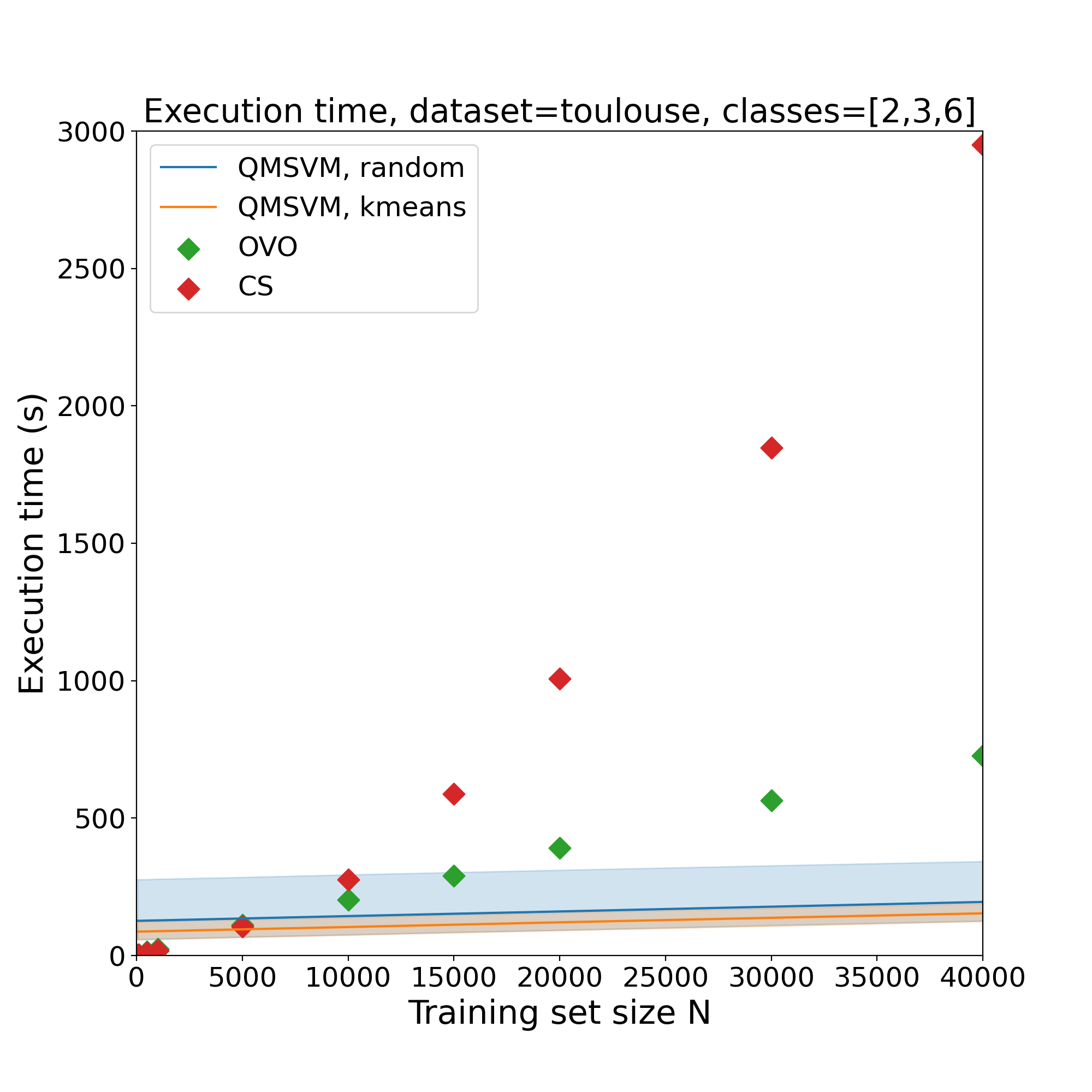}}
    \vspace{2mm}
    \caption{Toulouse - test accuracy and execution time for Quantum Multiclass SVM (QMSVM), one-versus-one (OVO) and Crammer-Singer SVM (CS) with respect to training set size $N$.}
    \label{fig:toulouse_accuracy_time}
\end{figure*}
%
\renewcommand\figurespace{0.5\linewidth}
\renewcommand\figurewidth{0.4\linewidth}
\begin{figure*}
    \centering
    \subfloat[Test accuracy vs. training set size N] {\includegraphics[width=\figurewidth]{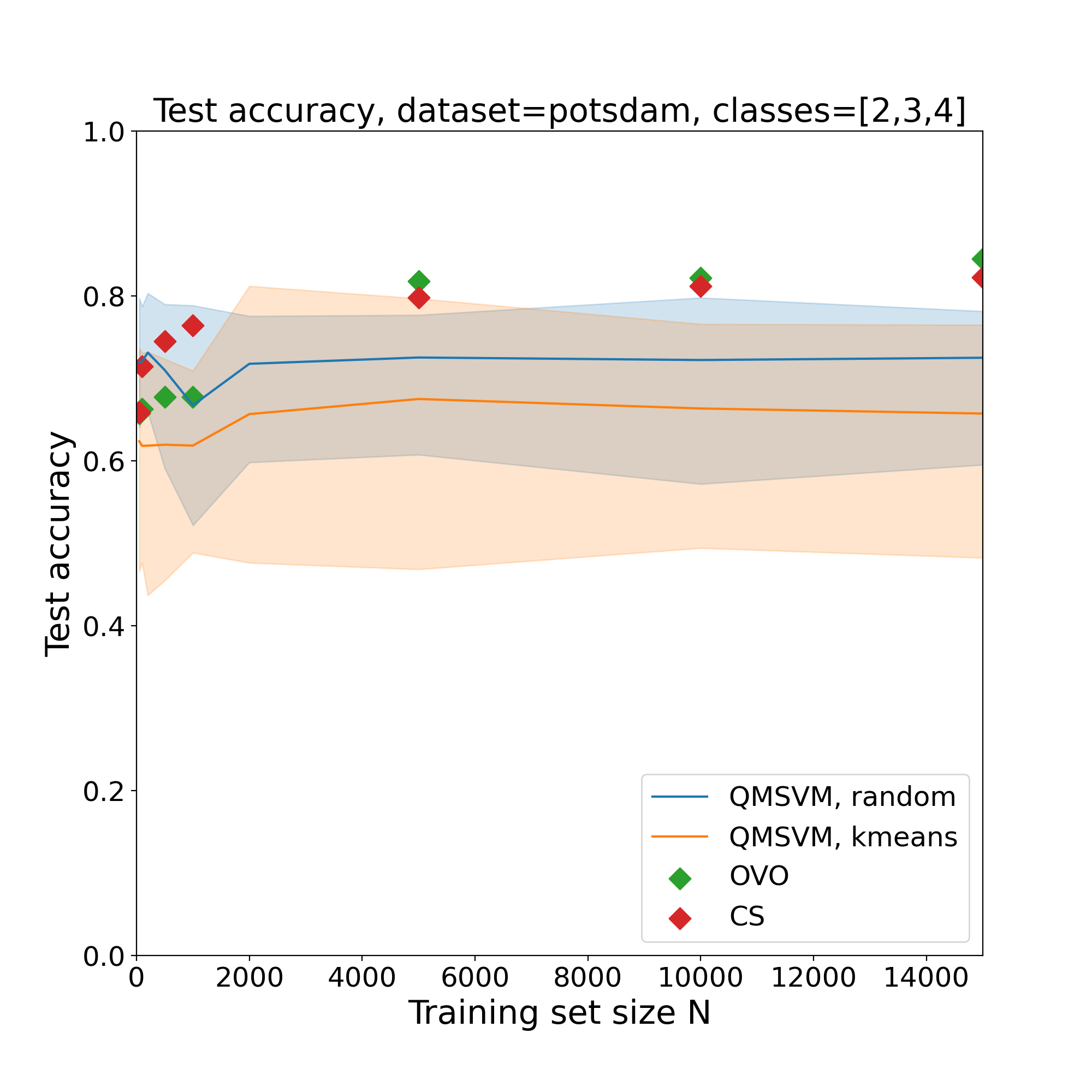}}
    \hfil
    \subfloat[Execution time vs. training set size N]{\includegraphics[width=\figurewidth]{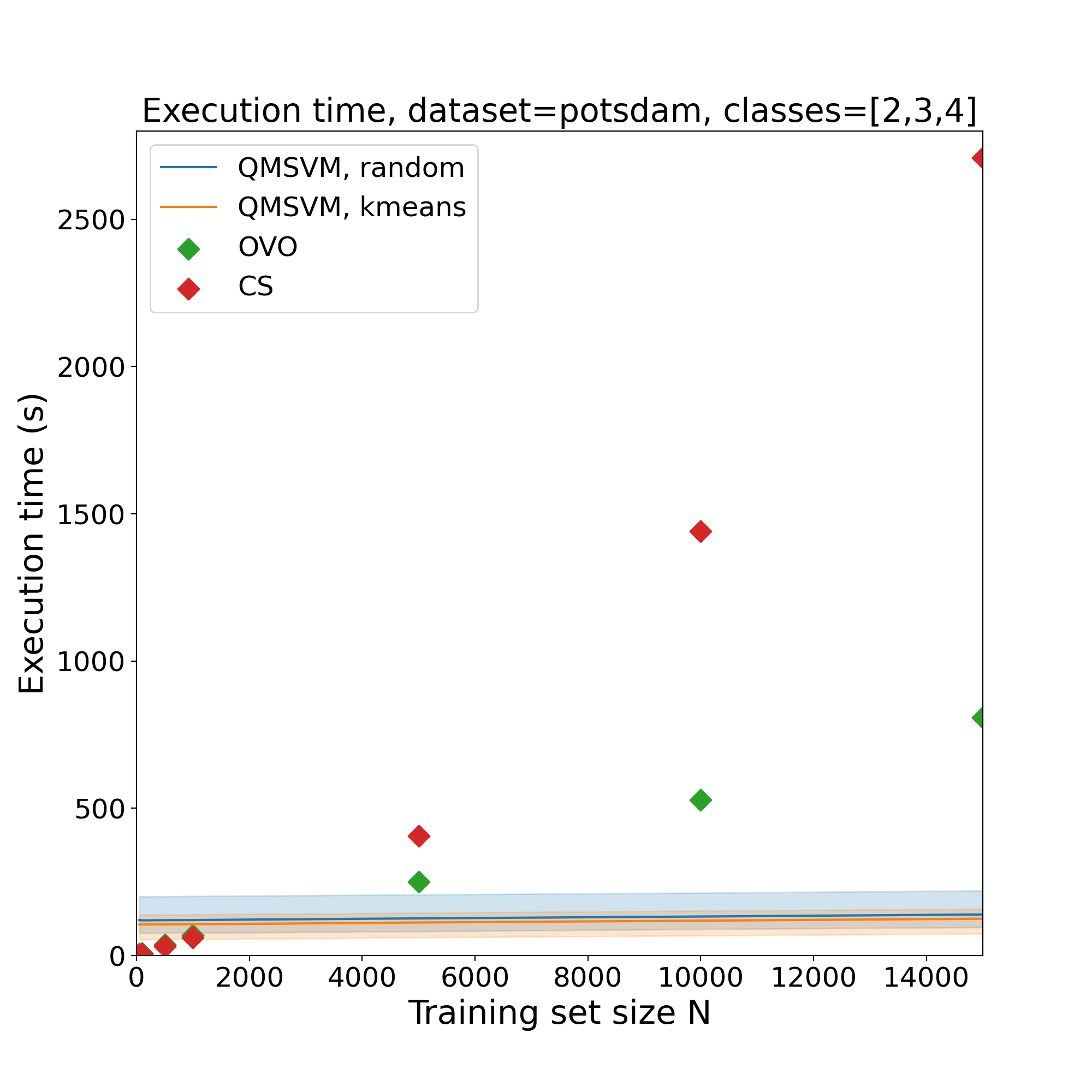}}
    \vspace{2mm}
    \caption{Potsdam - test accuracy and execution time for Quantum Multiclass SVM (QMSVM), one-versus-one (OVO) and Crammer-Singer SVM (CS) with respect to training set size $N$.}
    \label{fig:potsdam_accuracy_time}
\end{figure*}

\renewcommand\figurespace{0.5\linewidth}
\renewcommand\figurewidth{0.95\linewidth}
\begin{figure}
    \centering
    \subfloat[QMSVM (selection, annealing) - Execution time vs. training set size N] {\includegraphics[width=\figurewidth]{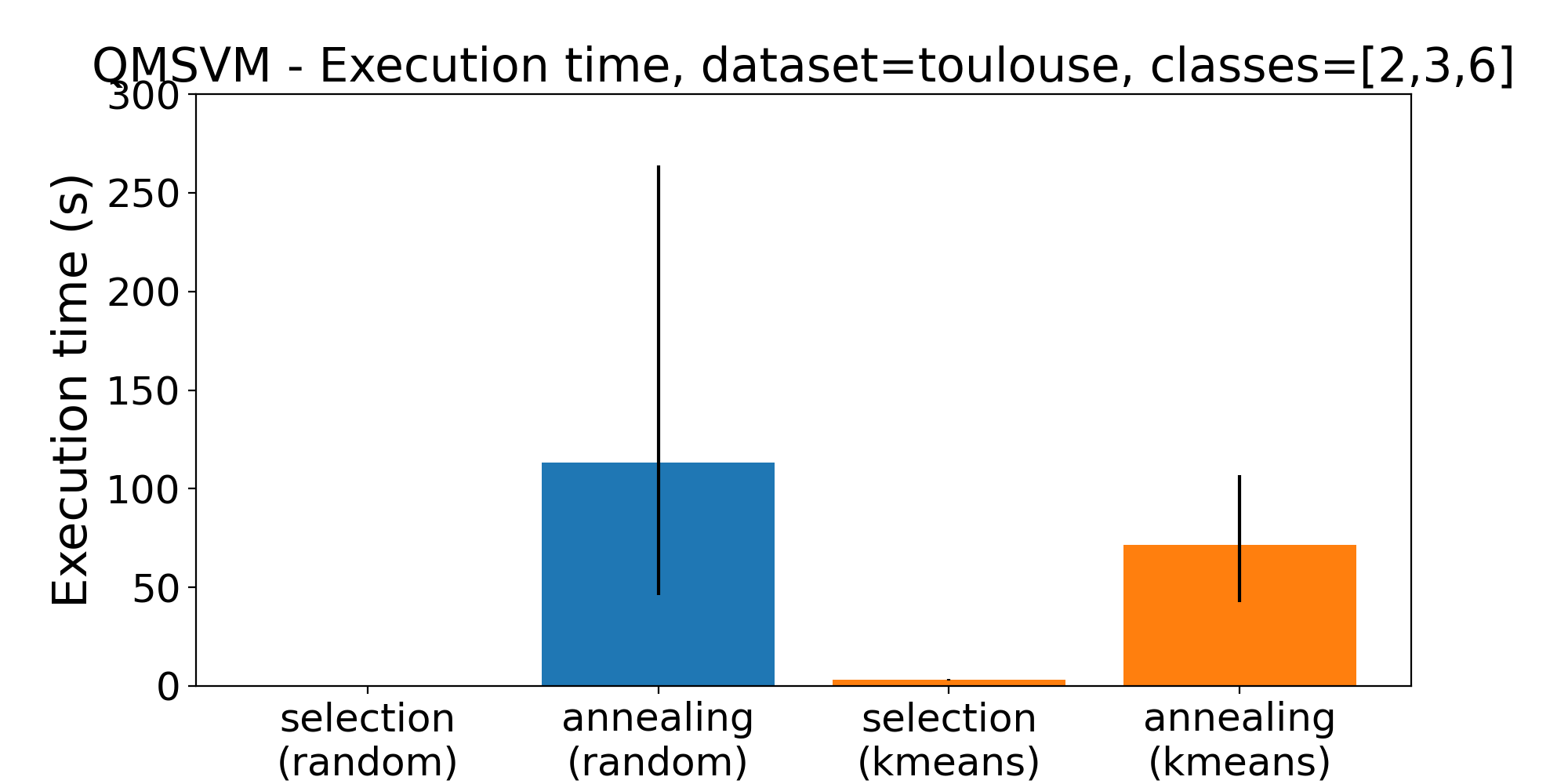}}\\
    \subfloat[QMSVM (combination, inference) - Execution time vs. training set size N]{\includegraphics[width=\figurewidth]{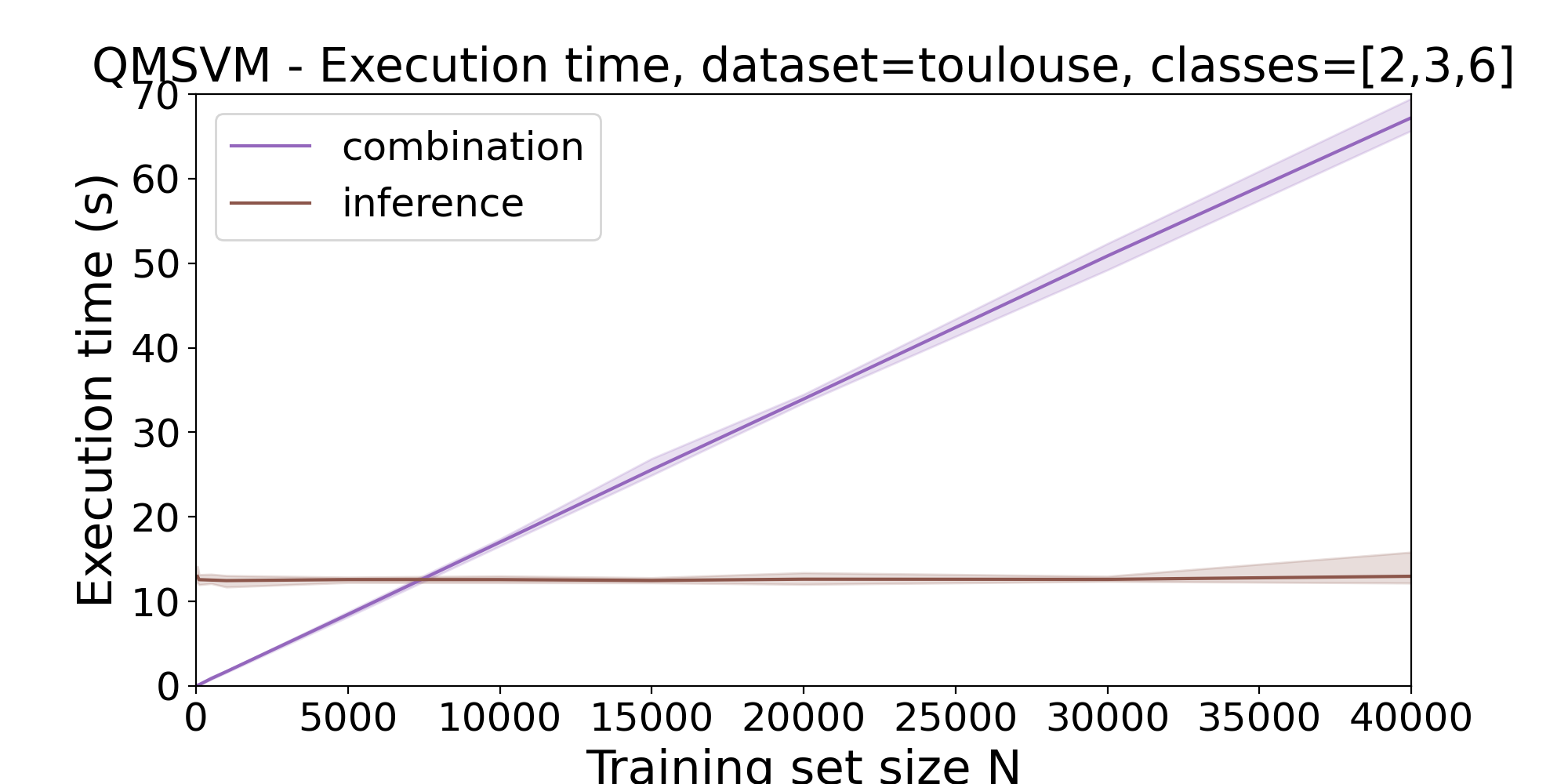}}\\
    \subfloat[OVO, CS - Execution time vs. training set size N]{\includegraphics[width=\figurewidth]{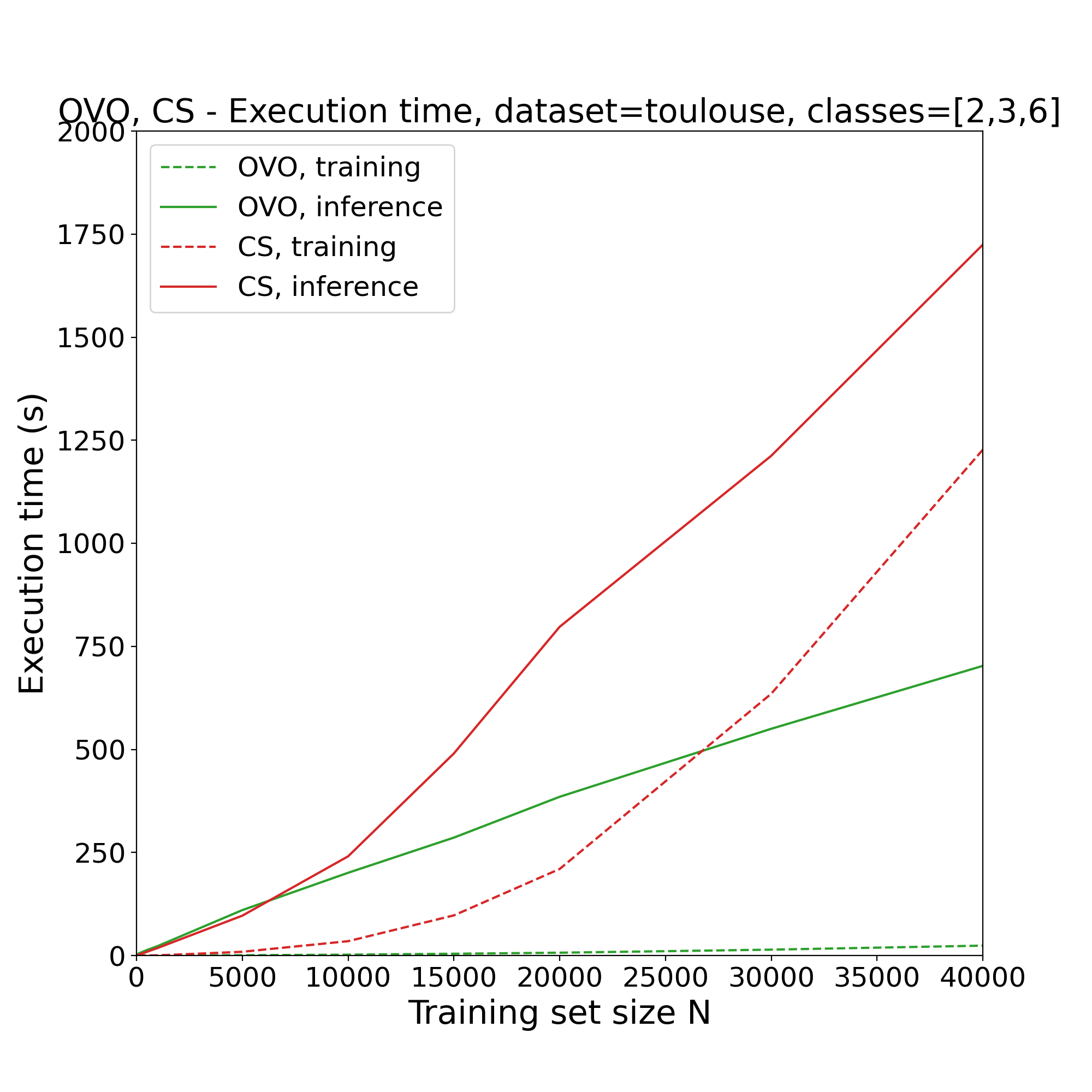}}
    \caption{Toulouse - execution time of each performed step, for Quantum Multiclass SVM (QMSVM), one-versus-one (OVO) and Crammer-Singer SVM (CS), with respect to training set size $N$.}
    \label{fig:toulouse_alltimes}
\end{figure}

\renewcommand\figurespace{0.5\linewidth}
\renewcommand\figurewidth{0.95\linewidth}
\begin{figure}
    \centering
    \subfloat[QMSVM (selection, annealing) - Execution time vs. training set size N] {\includegraphics[width=\figurewidth]{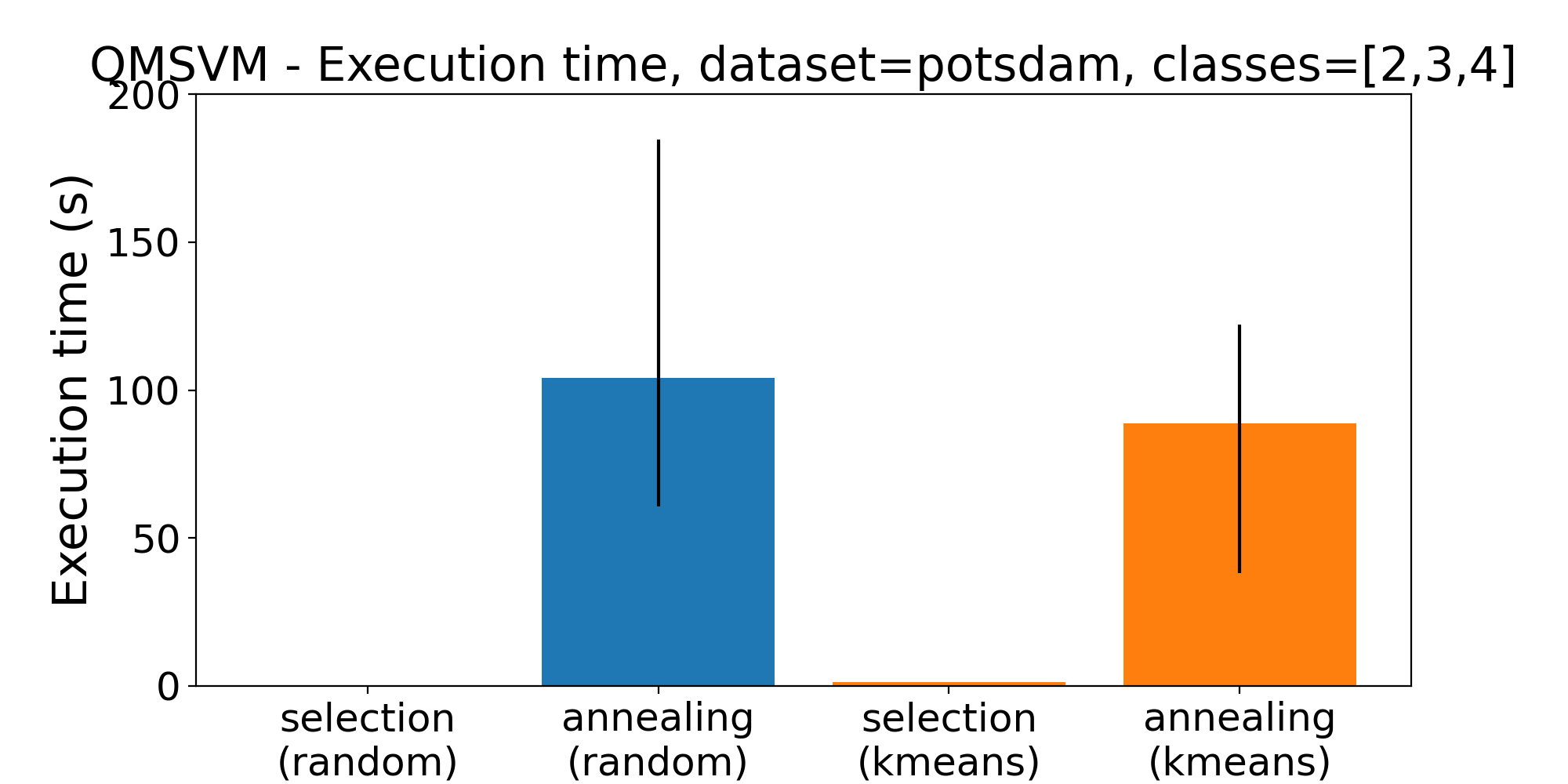}}\\
    \subfloat[QMSVM (combination, inference) - Execution time vs. training set size N] {\includegraphics[width=\figurewidth]{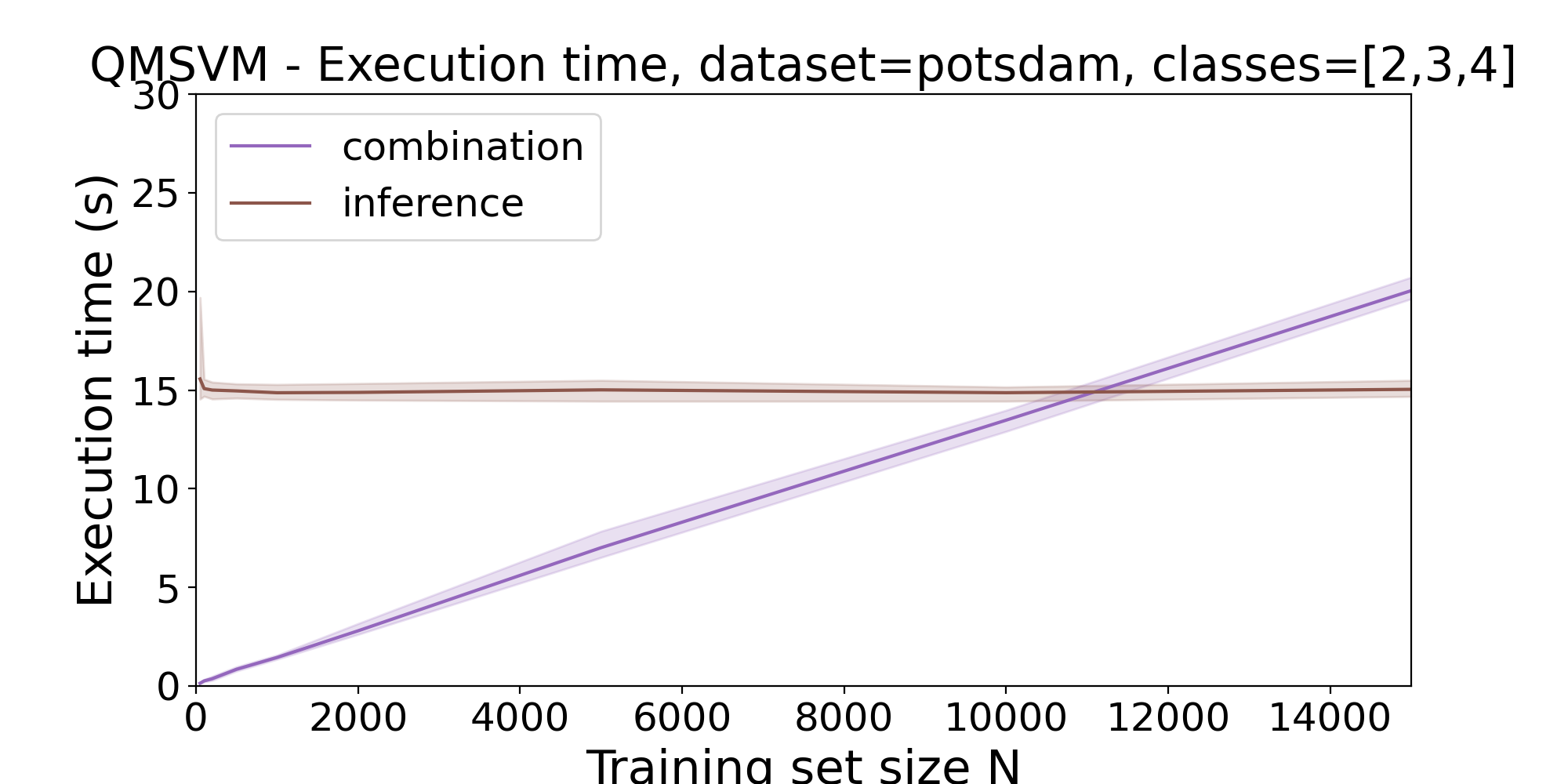}}\\
    \subfloat[OVO, CS - Execution time vs. training set size N] {\includegraphics[width=\figurewidth]{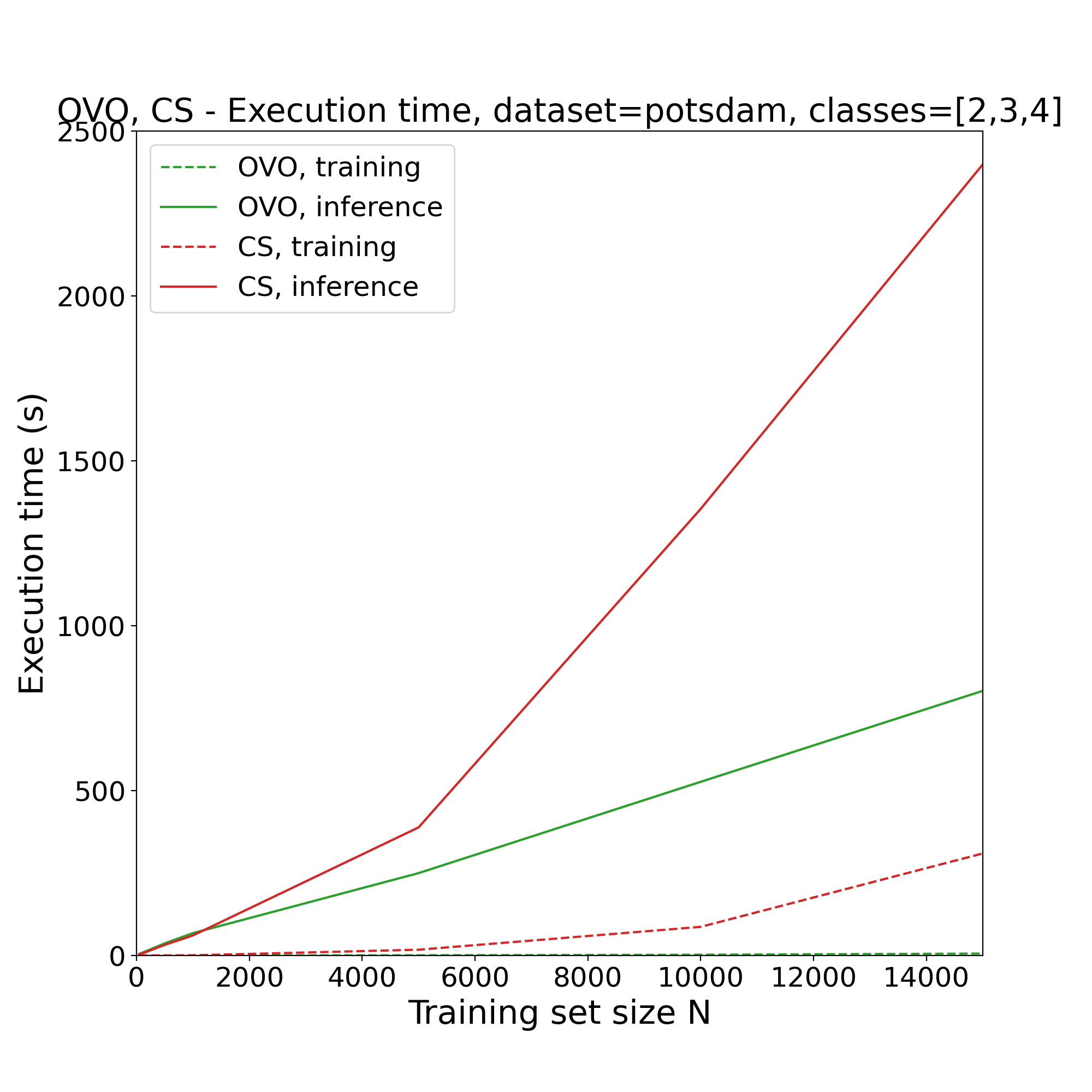}}
    \caption{Potsdam - execution time of each performed step, for Quantum Multiclass SVM (QMSVM), one-versus-one (OVO) and Crammer-Singer SVM (CS), with respect to training set size $N$.}
    \label{fig:potsdam_alltimes}
\end{figure}

\section{Conclusions}
\label{sec:conclusions}
\ac{QMSVM} serves as a preliminary framework for applying \ac{QA} to a single-step \ac{MSVM} algorithm, successfully leveraging the D-Wave Advantage quantum annealer in the training step. Although the results show that the prediction accuracy is not higher than standard \ac{MSVM} algorithms for the same training set, the improved scalability allows the usage of large-scale datasets.
Further research has to be conducted, in light of the promising achieved results. 
Time and accuracy analysis can be performed on different datasets, better assessing the impact of $N$ and the model parameters on the prediction accuracy. 
A deeper analysis of different selection methods based on dataset representativeness, on top of the k-means method, can improve the quality of the solutions obtained by the \ac{QA}.
An improvement in performance for \ac{QMSVM} is expected with the future development of \ac{QA}, as a higher memory and qubit connectivity allows the usage of a larger training subset and enhances the quality of the obtained solutions.

\section*{Acknowledgment}

The authors gratefully acknowledge support from the project JUNIQ that has received funding from the German Federal Ministry of Education and Research (BMBF) and the Ministry of Culture and Science of the State of North Rhine-Westphalia. This work is part of the Quantum Computing for Earth Observation (QC4EO) initiative from the ESA \textPhi-lab and the Center of Excellence (CoE) Research on AI- and Simulation-Based Engineering at Exascale (RAISE) receiving funding from EU’s Horizon 2020 Research and Innovation Framework Programme H2020-INFRAEDI-2019-1 under grant agreement no. 951733.
Icelandic HPC National Competence Center is funded by the EuroCC project no. 151454 that has received funding from the EU HPC Joint Undertaking (JU) under grant agreement no. 951732.}
\ifCLASSOPTIONcaptionsoff
  \newpage
\fi



%


\bibliographystyle{IEEEtran}
\bibliography{refs}

%

\begin{IEEEbiography}
[{\includegraphics[width=1in,height=1.25in,clip,keepaspectratio]{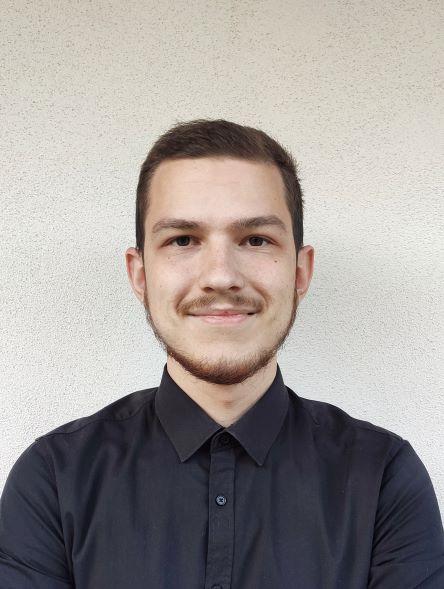}}]
{Amer Delilbasic} (Student Member, IEEE) received the B.Sc. and M.Sc. degrees in information and communication engineering from the University of Trento in 2019 and 2021, respectively.
He is member of the ``AI and ML for Remote Sensing'' Simulation and Data Lab at the J\"{u}lich Supercomputing Centre, Forschungszentrum J\"{u}lich, Germany. He is currently pursuing the Ph.D. degree in computational engineering at the University of Iceland. He is an external researcher at Φ-lab, European Space Agency, Frascati, Italy. His research interest is mainly in machine learning methods for remote sensing applications, with a particular focus on Quantum Computing (QC) and High Performance Computing (HPC).
\end{IEEEbiography}

\begin{IEEEbiography}[{\includegraphics[width=1in,height=1.25in,clip,keepaspectratio]{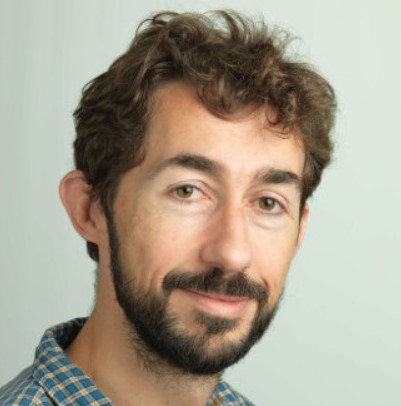}}]{Bertrand Le Saux} (Senior Member, IEEE) received the
Ms.Eng. and M.Sc. degrees from INP, Grenoble,
France, in 1999, the Ph.D. degree from the Univer-
sity of Versailles/Inria, Versailles, France, in 2003,
and the Dr. Habil. degree from the University of
Paris-Saclay, Saclay, France, in 2019. He is a Senior
Scientist with the European Space Agency/European
Space Research Institute \textPhi-lab in Frascati, Italy.
His research interest aims at visual understanding of
the environment by data-driven techniques includ-
ing Artificial Intelligence and (Quantum) Machine
Learning. He is interested in tackling practical problems that arise in Earth
observation, to bring solutions to current environment and population chal-
lenges. Dr. Le Saux is an Associate Editor of the Geoscience and Remote
Sensing Letters. He was Co-Chair (2015–2017) and chair (2017–2019) for
the IEEE GRSS Technical Committee on Image Analysis and Data Fusion.
\end{IEEEbiography}

\begin{IEEEbiography}[{\includegraphics[width=1in,height=1.25in,clip,keepaspectratio]{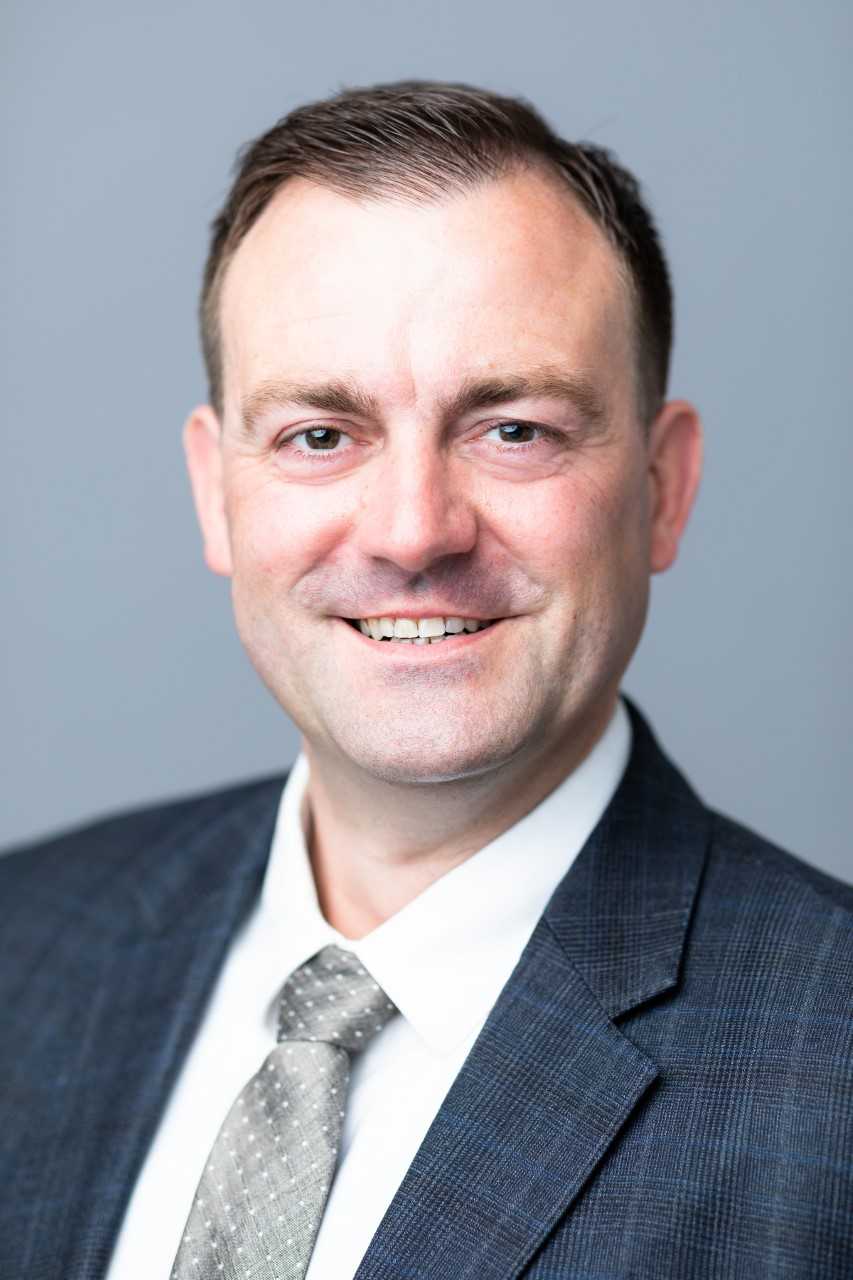}}]{Morris Riedel} (Member, IEEE) received his PhD from the Karlsruhe Institute of Technology (KIT) and worked in data-intensive parallel and distributed systems since 2004. He is currently a Full Professor of High-Performance Computing with an emphasis on Parallel and Scalable Machine Learning at the School of Natural Sciences and Engineering of the University of Iceland. Since 2004, Prof. Dr. - Ing. Morris Riedel held various positions at the Juelich Supercomputing Centre of Forschungszentrum Juelich in Germany. In addition, he is the Head of the joint High Productivity Data Processing research group between the Juelich Supercomputing Centre and the University of Iceland. Since 2020, he is also the EuroHPC Joint Undertaking governing board member for Iceland. His research interests include high-performance computing, remote sensing applications, medicine and health applications, pattern recognition, image processing, and data sciences, and he has authored extensively in those fields. Prof. Dr. – Ing. Morris Riedel online YouTube and university lectures include High-Performance Computing – Advanced Scientific Computing, Cloud Computing and Big Data – Parallel and Scalable Machine and Deep Learning, as well as Statistical Data Mining. In addition, he has performed numerous hands-on training events in parallel and scalable machine and deep learning techniques on cutting-edge HPC systems.
\end{IEEEbiography}

\begin{IEEEbiography}[{\includegraphics[width=1in,height=1.25in,clip,keepaspectratio]{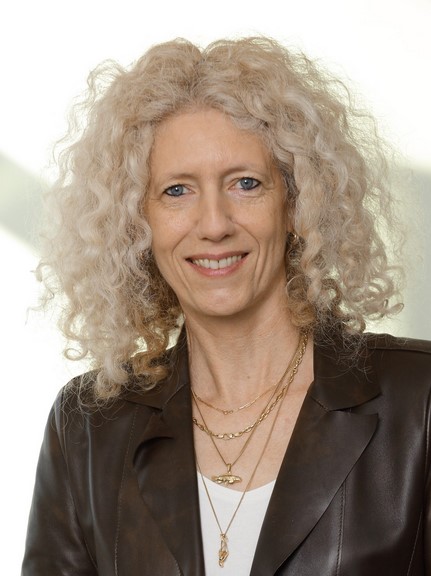}}]{Kristel Michielsen} received her PhD from the University of Groningen, the Netherlands, for work on the simulation of strongly correlated electron systems in 1993. Since 2009 she is group leader of the research group Quantum Information Processing at the Jülich Supercomputing Centre, Forschungszentrum Jülich (Germany) and is also Professor of Quantum Information Processing at RWTH Aachen University (Germany).
Her current research interests include quantum computation, quantum annealing, quantum statistical physics, event-based simulation methods of quantum phenomena, logical inference approach to quantum mechanics and computational electrodynamics.
\end{IEEEbiography}

\begin{IEEEbiography}
[{\includegraphics[width=1in,height=1.25in,clip,keepaspectratio]{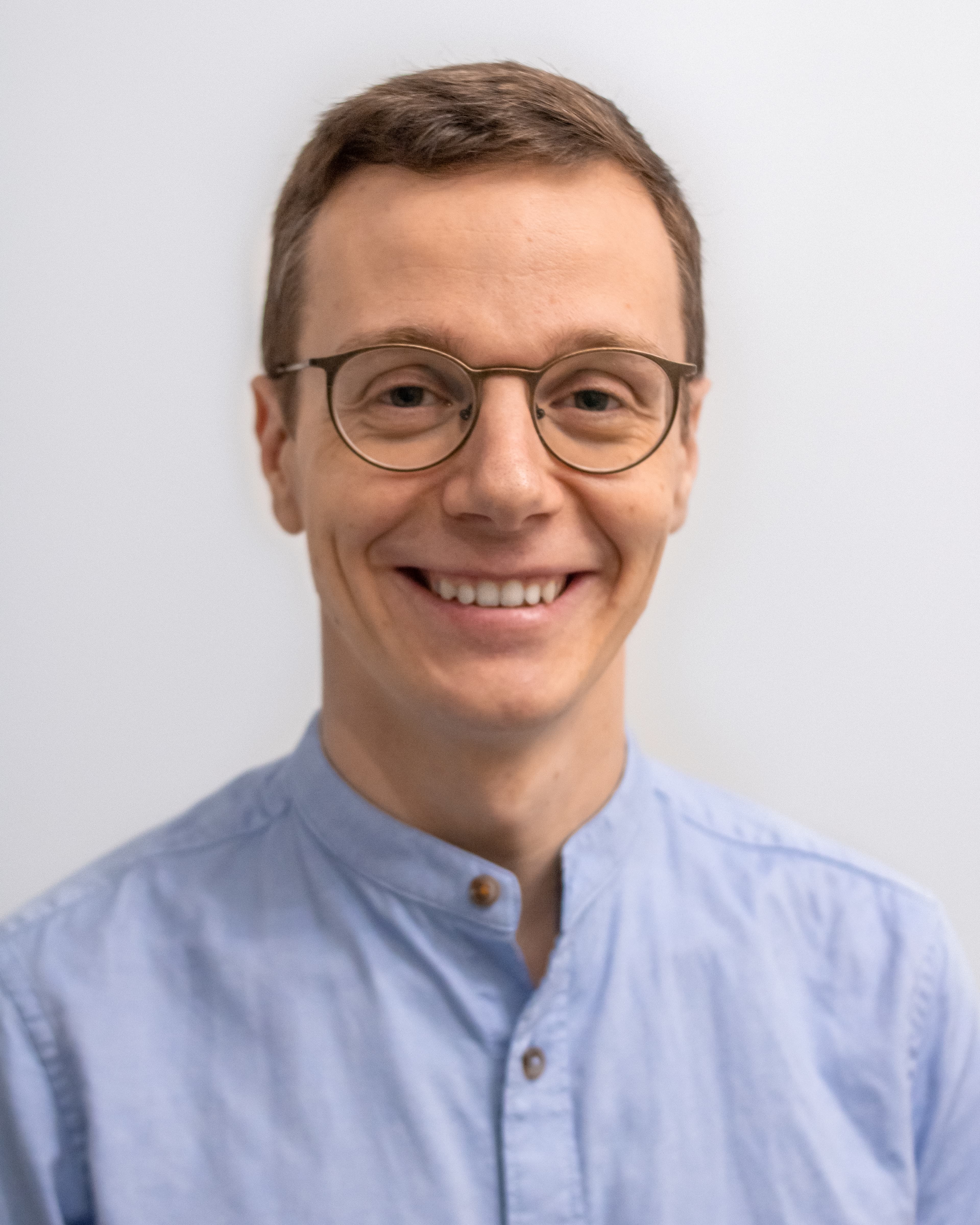}}]{Gabriele Cavallaro} (Member, IEEE) received his B.Sc. and M.Sc. degrees in Telecommunications Engineering from the University of Trento, Italy, in 2011 and 2013, respectively, and a Ph.D. degree in Electrical and Computer Engineering from the University of Iceland, Iceland, in 2016. From 2016 to 2021 he has been the deputy head of the ``High Productivity Data Processing'' (HPDP) research group at the J\"{u}lich Supercomputing Centre, Germany.  From 2019 to 2021 he gave lectures on scalable machine learning for remote sensing big data at the Institute of Geodesy and Geoinformation, University of Bonn, Germany. Since 2022, he is the Head of the ``AI and ML for Remote Sensing'' Simulation and Data Lab at the J\"{u}lich Supercomputing Centre, Forschungszentrum J\"{u}lich, Germany and an Adjunct Associate Professor with the School of Natural Sciences and Engineering, University of Iceland, Iceland. He is also the Chair of the High-Performance and Disruptive Computing in Remote Sensing (HDCRS) Working Group of the IEEE GRSS ESI Technical Committee and a Visiting Professor at the \textPhi-lab of the European Space Agency (ESA) in the context of the Quantum Computing for Earth Observation (QC4EO) initiative. Since October 2022 he serves as an Associate Editor of the IEEE Transactions on Image Processing (TIP). He also serves on the scientific committees of several international conferences and he is a referee for numerous international journals.

He was the recipient of the IEEE GRSS Third Prize in the Student Paper Competition of the IEEE International Geoscience and Remote Sensing Symposium (IGARSS) 2015 (Milan - Italy). His research interests cover remote sensing data processing with parallel machine learning algorithms that scale on distributed computing systems and cutting-edge computing technologies, including quantum computers. 

\end{IEEEbiography}




\end{document}